\documentclass[twoside]{article}

\usepackage[accepted]{aistats2e}
\usepackage{graphicx} 
\usepackage{subfigure} 
\usepackage{times}
\usepackage{epsfig}
\usepackage{graphicx}
\usepackage{amsmath}
\usepackage{amssymb}
\usepackage{subfigure}
\usepackage{cancel}
\usepackage{bm}
\usepackage{bbm}
\usepackage{nccmath}
\usepackage{natbib}
\usepackage[normalem]{ulem}
\usepackage{todonotes}
\usepackage[bottom]{footmisc}

\newcommand{\bfy}{\mathbf{y}}
\newcommand{\bfx}{\mathbf{x}}

\newcommand{\bfzi}{\mathbf{z}}
\newcommand{\bfzero}{\mathbf{0}}
\newcommand{\bfmu}{\boldsymbol \mu}

\newcommand{\bftheta}{\boldsymbol \theta}

\newcommand{\bfepsilon}{\boldsymbol \epsilon}
\newcommand{\bfF}{\mathbf{F}}

\newcommand{\bfI}{\mathbf{I}}
\newcommand{\bfK}{\mathbf{K}}
\newcommand{\bfg}{\mathbf{g}}
\newcommand{\bfr}{\mathbf{r}}

\newcommand{\T}{{\top}}

\newcommand{\ie}{i.e.\ }
\newcommand{\eg}{e.g.\ }

\newcommand{\intd}{\text{d}}

\global\long\def\dataScalar{y}
\global\long\def\dataMatrix{{\bf \MakeUppercase{\dataScalar}}}

\global\long\def\inputScalar{x}
\global\long\def\inputMatrix{{\bf \MakeUppercase{\inputScalar}}}
\global\long\def\inputVector{{\bf \inputScalar}}

\global\long\def\kernelMatrix{\mathbf{\MakeUppercase{\kernelScalar}}}
\global\long\def\kernelScalar{k}

\global\long\def\dataStd{\sigma}

\global\long\def\bfmu{\bm{\mu}}

\global\long\def\bfepsilon{\boldsymbol{\epsilon}}

\global\long\def\bftheta{\boldsymbol{\theta}}

\global\long\def\bff{\mathbf{f}}
\global\long\def\bfg{\mathbf{g}}

\global\long\def\bfr{\mathbf{r}}

\global\long\def\bfx{\mathbf{x}}
\global\long\def\bfy{\mathbf{y}}

\global\long\def\bfzero{\mathbf{0}}

\global\long\def\bfI{\mathbf{I}}

\global\long\def\bfK{\mathbf{K}}

\global\long\def\la{\leftarrow}

\global\long\def\T{{\rm T}}

\global\long\def\cut#1{}

\global\long\def\detail#1{}

\usepackage{color}
\usepackage{verbatim}

\global\long\def\dataStd{\sigma}

\global\long\def\dataScalar{y}

\global\long\def\dataMatrix{\mathbf{\MakeUppercase{\dataScalar}}}

\global\long\def\latentScalar{x}
\global\long\def\latentMatrix{\mathbf{\MakeUppercase{\latentScalar}}}

\global\long\def\kernelScalar{k}

\global\long\def\kernelMatrix{\mathbf{\MakeUppercase{\kernelScalar}}}

\global\long\def\mappingScalar{w}
\global\long\def\mappingVector{\mathbf{\mappingScalar}}

\global\long\def\mappingFunction{f}

\global\long\def\mappingFunctionMatrix{\mathbf{\MakeUppercase{\mappingFunction}}}

\global\long\def\gaussianDist#1#2#3{\mathcal{N}\left(#1|#2,#3\right)}

\global\long\def\la{\left\langle}
\global\long\def\ra{\right\rangle}

\begin{document}

%

%

\twocolumn[

\aistatstitle{Deep Gaussian Processes}


\aistatsauthor{Andreas C. Damianou \And  Neil D. Lawrence}
\aistatsaddress{Dept. of Computer Science \& Sheffield Institute for Translational Neuroscience, \\
University of Sheffield, UK}]

\begin{abstract}
  In this paper we introduce deep Gaussian process (GP) models. Deep
  GPs are a deep belief network based on Gaussian process
  mappings. The data is modeled as the output of a multivariate GP. The
  inputs to that Gaussian process are then governed by another GP. A
  single layer model is equivalent to a standard GP or the GP latent
  variable model (GP-LVM). We perform inference in the model by
  approximate variational marginalization. This results in a strict
  lower bound on the marginal likelihood of the model which we use for model
  selection (number of layers and nodes per layer). Deep belief
  networks are typically applied to relatively large data sets using
  stochastic gradient descent for optimization. Our fully Bayesian
  treatment allows for the application of deep models even when data
  is scarce. Model selection by our variational bound shows that a
  five layer hierarchy is justified even when modelling a digit data
  set containing only 150 examples.
\end{abstract}
\section{Introduction}

Probabilistic modelling with neural network architectures constitute a
well studied area of machine learning. The recent advances in the
domain of deep learning \citep{Hinton06afast,DeepLearningReview} have
brought this kind of models again in popularity. Empirically, deep
models seem to have structural advantages that can improve the quality
of learning in complicated data sets associated with abstract
information \citep{Bengio:DeepArchForAI}. Most deep algorithms require
a large amount of data to perform learning, however, we know that
humans are able to perform inductive reasoning (equivalent to concept
generalization) with only a few examples
\citep{Tenenbaum:theory06}. This provokes the question as to whether
deep structures and the learning of abstract structure can be undertaken in
\emph{smaller} data sets. For smaller data sets, questions of generalization 
arise: to demonstrate such structures are justified it is useful to have an 
objective measure of the model's applicability.

The traditional approach to deep learning is based around binary
latent variables and the restricted Boltzmann machine (RBM)
\citep{rbm}. Deep hierarchies are constructed by stacking
these models and various approximate inference techniques (such as
contrastive divergence) are used for estimating model parameters. 
A significant amount of work has then to be done with annealed
importance sampling if even the \emph{likelihood}\footnote{We use
  emphasis to clarify we are referring to the model likelihood, not
  the marginal likelihood required in Bayesian model selection.} of a
data set under the RBM model is to be estimated
\citep{Salakhutdinov:quantitative08}. When deeper hierarchies are
considered, the estimate is only of a lower bound on the data
likelihood. Fitting such models to smaller data sets and using
Bayesian approaches to deal with the complexity seems completely
futile when faced with these intractabilities.

The emergence of the Boltzmann machine (BM) at the core of one of the most
interesting approaches to modern machine learning is very much a case
of a the field going back to the future: BMs rose to
prominence in the early 1980s, but the practical implications
associated with their training led to their neglect until families of
algorithms were developed for the RBM model with its reintroduction as
a product of experts in the late nineties
\citep{Hinton:product99}. 

The computational intractabilities of Boltzmann machines led to other
families of methods, in particular kernel methods such as the support
vector machine (SVM), to be considered for the domain of
data classification. 
%
Almost contemporaneously to the SVM, Gaussian process (GP) models
\citep{Rasmussen:book06} were introduced as a fully probabilistic
substitute for the multilayer perceptron (MLP), inspired by the
observation \citep{Neal:book96} that, under certain conditions, a GP \emph{is} an MLP 
with \emph{infinite} units in the hidden layer. MLPs
also relate to deep learning models: deep learning
algorithms have been used to pretrain autoencoders for dimensionality
reduction \citep{Hinton:reducing06}.
%
Traditional GP models have been extended to more expressive variants, for example by considering
sophisticated covariance functions \citep{Durrande:additive, Gonen:multiple} 
or by embedding GPs in more complex probabilistic structures \citep{snelson:warpedgaussian, wilson:gprn} 
able to learn more powerful representations
of the data. However, all GP-based approaches considered so far do not lead
to a principled way of obtaining truly deep architectures and, to date, the field of deep learning
remains mainly associated with RBM-based models.

The conditional probability of a single hidden unit in an RBM model, given its parents, is written as
\[
p(\dataScalar|\inputVector) = \sigma(\mappingVector^\top \inputVector)^{\dataScalar}(1-\sigma(\mappingVector^\top\inputVector))^{(1-\dataScalar)},
\]
where here $\dataScalar$ is the output variable of the RBM,
$\inputVector$ is the set of inputs being conditioned on and
$\sigma(z)=(1+\exp(-z))^{-1}$. The conditional density of the output
depends only on a linear weighted sum of the inputs. The
representational power of a Gaussian process in the same role is
significantly greater than that of an RBM. For the GP the
corresponding likelihood is over a continuous variable, but it is a
nonlinear function of the inputs,
\[
p(\dataScalar|\inputVector) = \gaussianDist{\dataScalar}{\mappingFunction(\inputVector)}{\sigma^2},
\]
where $\gaussianDist{\cdot}{\mu}{\sigma^2}$ is a Gaussian density with
mean $\mu$ and variance $\sigma^2$. In this case the likelihood is
dependent on a mapping function, $\mappingFunction(\cdot)$, rather
than a set of intermediate parameters, $\mappingVector$. The approach
in Gaussian process modelling is to place a prior directly over the
classes of functions (which often specifies smooth, stationary
nonlinear functions) and integrate them out. This can be done
analytically. In the RBM the model likelihood is estimated and
maximized with respect to the parameters, $\mappingVector$. For the
RBM marginalizing $\mappingVector$ is not analytically tractable. We
note in passing that the two approaches can be mixed if
$
p(\dataScalar|\inputVector) = \sigma(\mappingFunction(\inputVector))^{\dataScalar}(1-\sigma(\mappingFunction(\inputVector))^{(1-\dataScalar)},
$
which recovers a GP classification model. Analytic integration is no
longer possible though, and a common approach to approximate inference
is the expectation propagation algorithm \citep[see
e.g.][]{Rasmussen:book06}. However, we don't consider this idea
further in this paper.

Inference in deep models requires marginalization of $\inputVector$ as
they are typically treated as \emph{latent} variables\footnote{They
  can also be treated as observed, \eg in the upper most layer
  of the hierarchy where we might include the data label.},
which in the case of the RBM are binary
variables. The number of the terms in the sum scales exponentially
with the input dimension rendering it intractable for anything but the
smallest models. In practice, sampling and, in particular, the
contrastive divergence algorithm, are used for training. Similarly,
marginalizing $\inputVector$ in the GP is analytically intractable,
even for simple prior densities like the Gaussian. In the GP-LVM
\citep{Lawrence:pnpca05} this problem is solved through maximizing
with respect to the variables (instead of the parameters, which are
marginalized) and these models have been combined in stacks to form
the hierarchical GP-LVM \citep{Lawrence:hgplvm07} which is a maximum a
posteriori (MAP) approach for learning deep GP models. For this MAP
approach to work, however, a strong prior is required on the top level
of the hierarchy to ensure the algorithm works and MAP learning
prohibits model selection because no estimate of the marginal
likelihood is available.

There are two main contributions in this paper. Firstly, we exploit
recent advances in variational inference \citep{Titsias:bayesGPLVM10}
to marginalize the latent variables in the hierarchy
variationally. \citet{Damianou:vgpds11} has already shown how using
these approaches two Gaussian process models can be stacked.
This paper goes further to show that through variational approximations any
number of GP models can be stacked to give truly deep hierarchies. The
variational approach gives us a rigorous lower bound on the
\emph{marginal} likelihood of the model, allowing it to be used for
model selection. 
Our second contribution is to use this lower bound to demonstrate the
applicability of deep models even when data is scarce. The variational
lower bound gives us an objective measure from which we can select
different structures for our deep hierarchy (number of layers, number
of nodes per layer). In a simple digits example we find that the best
lower bound is given by the model with the deepest hierarchy we
applied (5 layers).

The deep GP consists of a cascade of hidden layers of latent variables
where each node acts as output for the layer above and as input for
the layer below---with the observed outputs being placed in the
leaves of the hierarchy. Gaussian processes govern the mappings
between the layers.

A single layer of the deep GP is effectively a Gaussian process latent
variable model (GP-LVM), just as a single layer of a regular deep
model is typically an RBM.  \citep{Titsias:bayesGPLVM10} have shown that
latent variables can be approximately marginalized in the GP-LVM
allowing a variational lower bound on the likelihood to be
computed. The appropriate size of the latent space can be computed
using automatic relevance determination (ARD) priors
\citep{Neal:book96}.
 \cite{Damianou:vgpds11}
extended this approach by placing a GP prior over the
latent space, resulting in a Bayesian dynamical GP-LVM. Here we extend that
approach to allow us to approximately marginalize any number of hidden
layers. We demonstrate how a deep hierarchy of Gaussian processes can
be obtained by marginalising out the latent variables in the
structure, obtaining an approximation to the fully Bayesian training
procedure and a variational approximation to the true posterior
of the latent variables given the outputs. The resulting model is
very flexible and should open up a range of applications for deep structures
\footnote{A preliminary version of this paper has been presented in \citep{Damianou:deepGPsWorkshop}.}.





\section{The Model}

We first consider standard approaches to modeling with GPs. We then
extend these ideas to deep GPs by considering Gaussian process priors
over the inputs to the GP model. We can apply this idea recursively to
obtain a deep GP model.

\subsection{Standard GP Modelling}

In the traditional probabilistic inference framework, we are given a
set of training input-output pairs, stored in matrices $\inputMatrix \in
\mathcal{R}^{N \times Q}$ and $\dataMatrix \in \mathcal{R}^{N \times D}$
respectively, and seek to estimate the unobserved, \textit{latent}
function $f=f(\inputVector)$, responsible for generating $\dataMatrix$ given $\inputMatrix$. 
In this setting, Gaussian processes (GPs) \citep{Rasmussen:book06} can
be employed as nonparametric prior distributions over the latent
function $f$. More formally, we assume that each datapoint $\bfy_n$ is
generated from the corresponding $f(\bfx_n)$ by adding independent
Gaussian noise, \ie
\begin{equation}
\label{generativeProcessGP}
\bfy_{n} = f(\bfx_n) + \epsilon_{n} , \; \; \epsilon \sim \mathcal{N}(\bfzero, \sigma_{\epsilon}\bfI),
\end{equation}
and $f$ is drawn from a Gaussian process, \ie $f(\bfx) \sim
\mathcal{GP}\left(\bfzero, k(x, x') \right)$.  This (zero-mean)
Gaussian process prior only depends on the covariance function $k$
operating on the inputs
$\inputMatrix$. 
As we wish to obtain a flexible model, we only make very general
assumptions about the form of the generative mapping $f$ and this is
reflected in the choice of the covariance function which defines the
properties of this mapping.
For example, an exponentiated quadratic covariance function,
$
 k \left( \bfx_i, \bfx_j \right) =  
(\sigma_{se})^2 \exp \left( - \frac{\left( \bfx_i - \bfx_j \right)^2}{2 l^2} \right)
$,
forces the latent functions to be infinitely smooth. We denote
any covariance function hyperparameters (such as $(\sigma_{se}, l)$ of
the aforementioned covariance function) by $\bftheta$.  The collection of latent
function instantiations, denoted by $\bfF = \{\bff_n\}_{n}^N$, is
normally distributed, allowing us to compute analytically the marginal
likelihood \footnote{All probabilities involving $f$ should also have
  $\bftheta$ in the conditioning set, but here we omit it for
  clarity.}
\begin{align}
\label{GPmarginal}
&p(\dataMatrix | \inputMatrix) = \int \prod_{n=1}^N p(\bfy_n | \bff_n) p(\bff_n | \bfx_n) \intd \bfF \nonumber \\
&= \mathcal{N}(\dataMatrix | \bfzero, \kernelMatrix_{NN} + \dataStd_{\epsilon}^2\bfI), \kernelMatrix_{NN}=k(\inputMatrix,\inputMatrix) .
\end{align}

Gaussian processes have also been used with success in unsupervised
learning scenarios, where the input data $\latentMatrix$ are not directly
observed.  The Gaussian process latent variable model (GP-LVM)
\citep{Lawrence:pnpca05,GPLVM2} provides an elegant solution to this problem by
treating the unobserved inputs $\latentMatrix$ as latent variables, while
employing a product of $D$ independent GPs as prior for the latent
mapping. The assumed generative procedure takes the form:
$y_{nd} = f_d(\bfx_n) + \epsilon_{nd},$
where $\bfepsilon$ is again Gaussian with variance $\dataStd_\epsilon^2$ and $\bfF =
\{\bff_d\}_{d=1}^D$ with $f_{nd} = f_d(\bfx_n)$.  Given a finite
data set, the Gaussian process priors take the form
\begin{equation}
\label{GppriorDistribution}
p(\mappingFunctionMatrix | \inputMatrix) = \prod_{d=1}^D \mathcal{N}(\bff_d | \bfzero, \bfK_{NN})
\end{equation}
which is a Gaussian and, thus, allows for general non-linear mappings to be
marginalised out analytically to obtain the likelihood $p(\dataMatrix|\inputMatrix) =
\prod_{d=1}^D \mathcal{N}(\bfy_d | \bfzero, \kernelMatrix_{NN}+\dataStd_\epsilon^2)$, analogously to equation
\eqref{GPmarginal}.


\subsection{Deep Gaussian Processes}

Our deep Gaussian process architecture corresponds to a graphical
model with three kinds of nodes, illustrated in figure
\ref{fig:grModel}\subref{hierarchy}: the leaf nodes $\dataMatrix \in
\mathcal{R}^{N \times D}$ which are observed, the
intermediate latent spaces $\latentMatrix_h \in \mathcal{R}^{N \times Q_h},
h=1,..., H-1$, where $H$ is the number of hidden layers, and the
parent latent node $\mathbf{Z} = \latentMatrix_H \in \mathcal{R}^{N \times Q_Z}$.
The parent node can be unobserved and potentially constrained with a prior
of our choice (\eg a dynamical prior), 
or could constitute the given inputs for a supervised learning task.
For simplicity, here we focus on the unsupervised learning scenario.
%
In this deep architecture, all intermediate
nodes $\latentMatrix_h$ act as inputs for the layer below (including the leaves)
and as outputs for the layer above. For simplicity, consider a
structure with only two hidden units, as the one depicted in figure
\ref{fig:grModel}\subref{hierarchySimple}. The generative process
takes the form:
\begin{align}
 y_{nd} = & f^Y_d(\bfx_n) + \epsilon^Y_{nd}, \; \;   d=1,..., D , \;\;   \bfx_n \in \mathcal{R}^Q \nonumber \\
 x_{nq} = & f^X_q(\bfzi_n) + \epsilon^X_{nq}, \; \;   q=1,..., Q, \;\;   \bfzi_n \in \mathcal{R}^{Q_Z}
\end{align}
and the intermediate node is involved in two Gaussian processes, $f^Y$ and $f^X$, playing the role
of an input and an output respectively:
 $f^Y \sim \mathcal{GP}(\bfzero, k^Y(\latentMatrix, \latentMatrix))  \; \; \; \text{and} \; \; \; f^X  \sim \mathcal{GP}(\bfzero, k^X(\mathbf{Z}, \mathbf{Z}))$.
This structure can be naturally extended vertically (\ie deeper
hierarchies) or horizontally (\ie segmentation of each layer into
different partitions of the output space), as we will see later in the
paper. However, it is already obvious how each layer adds a
significant number of model parameters ($\latentMatrix_h$) as well as a
regularization challenge, since the size of each latent layer is
crucial but has to be a priori defined.  For this reason, unlike
\cite{Lawrence:hgplvm07}, we seek to variationally
marginalise out the whole latent space.  Not only this will allow us
to obtain an automatic Occam's razor due to the Bayesian training, but
also we will end up with a significantly lower number of model
parameters, since the variational procedure only adds variational
parameters. The first step to this approach is to define automatic
relevance determination (ARD) covariance functions for the GPs:
\begin{align}
k \left( \mathbf{x}_i, \mathbf{x}_j \right) = {} &  
		\sigma_{ard}^2 e^{
			- \frac{1}{2} \sum_{q=1}^{Q}  w_q \left(
                          \mathit{x_{i,q} - x_{j,q}} \right) ^2 }.
\label{rbfard}
\end{align}
This covariance function assumes a different weight $w_q$ for each
latent dimension and this can be exploited in a Bayesian training
framework in order to ``switch off'' irrelevant dimensions by driving
their corresponding weight to zero, thus helping towards automatically
finding the structure of complex models. 
However, the nonlinearities introduced by this covariance function
make the Bayesian treatment of this model challenging.
Nevertheless, following recent non-standard variational inference methods
we can define analytically an approximate Bayesian training procedure,
as will be explained in the next section.

\subsection{Bayesian Training}

A Bayesian training procedure requires optimisation of the model evidence:
\begin{equation}
\label{pY}
 \log p(\dataMatrix) =  \log \int_{\latentMatrix,\mathbf{Z}} p(\dataMatrix|\latentMatrix) p(\latentMatrix|\mathbf{Z}) p(\mathbf{Z}).
\end{equation}
When prior information is available regarding the observed data (\eg their dynamical nature is known a priori), the prior distribution on the parent latent node
can be selected so as to constrain the whole latent space through propagation of the prior density through the cascade. Here we take the general case where $p(\mathbf{Z}) = \mathcal{N}(\mathbf{Z} | \bfzero, I)$.
However, the integral of equation \eqref{pY} is intractable due to the nonlinear way in which $\latentMatrix$ and $\mathbf{Z}$ are treated through the GP priors $f^Y$ and $f^X$.
As a first step, we apply Jensen's inequality to find a variational lower bound 
$\mathcal{F}_v \le \log p(\dataMatrix)$, with
\begin{equation}
\label{jensens}
 \mathcal{F}_v = \int_{\latentMatrix, \mathbf{Z}, \mathbf{F}^Y, \mathbf{F}^X} \mathcal{Q}
  \log \frac{p(\dataMatrix,\mathbf{F}^Y,\mathbf{F}^X,\latentMatrix,\mathbf{Z})}{\mathcal{Q}} , 
\end{equation}
where we introduced a variational distribution $\mathcal{Q}$, the form of which
will be defined later on. By noticing that the joint distribution appearing above
can be expanded in the form
\begin{align}
\label{jointFull}
p(\dataMatrix, & \mathbf{F}^Y,\mathbf{F}^X,\latentMatrix,\mathbf{Z}) = \nonumber \\
         & p(\dataMatrix|\mathbf{F}^Y) p(\mathbf{F}^Y|\latentMatrix) p(\latentMatrix|\mathbf{F}^X) p(\mathbf{F}^X|\mathbf{Z}) p(\mathbf{Z}), 
\end{align}
we see that the integral of equation \eqref{jensens} is still intractable
because $\latentMatrix$ and $\mathbf{Z}$ still appear nonlinearly in the $p(\mathbf{F}^Y|\latentMatrix)$ and
$p(\mathbf{F}^X|\mathbf{Z})$ terms respectively.  A key result of \citep{Titsias:bayesGPLVM10}
is that expanding the probability space of the GP prior $p(\bfF|\latentMatrix)$ with
extra variables allows for priors on the latent space to be propagated
through the nonlinear mapping $f$. More precisely, we augment the
probability space of equation \eqref{GppriorDistribution} with $K$
auxiliary pseudo-inputs $\tilde{\latentMatrix} \in \mathcal{R}^{K \times Q}$ and
$\tilde{\mathbf{Z}} \in \mathcal{R}^{K \times Q_Z}$ that correspond to a
collection of function values $\mathbf{U}^Y \in \mathcal{R}^{K \times D}$ and
$\mathbf{U}^X \in \mathcal{R}^{K \times Q}$ respectively \footnote{The number
  of inducing points, $K$, does not need to be the same for every GP of the overall deep structure.}.
Following this approach, we obtain the augmented probability space: 
$p( \dataMatrix, \mathbf{F}^Y,\mathbf{F}^X,\latentMatrix,\mathbf{Z}, \mathbf{U}^Y, \mathbf{U}^X, \tilde{\latentMatrix}, \tilde{\mathbf{Z}}) = $
\begin{align}
       & p(\dataMatrix|\mathbf{F}^Y) p(\mathbf{F}^Y | \mathbf{U}^Y,\latentMatrix) p(\mathbf{U}^Y|\tilde{\latentMatrix})  \nonumber \\
 \cdot & p(\latentMatrix|\mathbf{F}^X) p(\mathbf{F}^X | \mathbf{U}^X,\mathbf{Z}) p(\mathbf{U}^X|\tilde{\latentMatrix}) p(\mathbf{Z})  \label{augmentedGPprior}
\end{align}
The pseudo-inputs $\tilde{\latentMatrix}$ and $\tilde{\mathbf{Z}}$ are known as \textit{inducing points},
and will be dropped from our expressions from now on, for clarity. 
Note that $\mathbf{F}^{Y}$ and $\mathbf{U}^Y$ are draws from the same GP so that 
 $p(\mathbf{U}^Y)$ and $p(\mathbf{F}^Y | \mathbf{U}^Y, \latentMatrix)$ are also Gaussian distributions
 (and similarly for $p(\mathbf{U}^X), p(\mathbf{F}^X|\mathbf{U}^X, \mathbf{Z})$).

We are now able to define a variational distribution $\mathcal{Q}$
which, when combined with the new expressions for the augmented GP
priors, results in a tractable variational bound. Specifically, we
have:
\begin{align}
 \mathcal{Q} = &    p(\mathbf{F}^Y|\mathbf{U}^Y,\latentMatrix) q(\mathbf{U}^Y) q(\latentMatrix) \nonumber \\
         \cdot & p(\mathbf{F}^X|\mathbf{U}^X,\mathbf{Z}) q(\mathbf{U}^X) q(\mathbf{Z}). \label{varDistr}
\end{align}

We select $q(\mathbf{U}^Y)$ and $q(\mathbf{U}^X)$ to be free-form variational distributions, while $q(\latentMatrix)$ and $q(\mathbf{Z})$ are chosen to be 
Gaussian, factorised with respect to dimensions:
\begin{equation}
  q(\latentMatrix)  = \prod_{q=1}^{Q} \mathcal{N}(\bfmu^X_q, \mathbf{S}^X_q), \; 
  q(\mathbf{Z})  = \prod_{q=1}^{Q_Z} \mathcal{N}(\bfmu^Z_q, \mathbf{S}^Z_q) .  \label{qX}
\end{equation}
By substituting equation \eqref{varDistr} back to \eqref{jensens} while also replacing the original joint distribution
with its augmented version in equation \eqref{augmentedGPprior}, we see that the
``difficult'' terms $p(\mathbf{F}^Y|\mathbf{U}^Y,\latentMatrix)$ and $p(\mathbf{F}^X|\mathbf{U}^X,\mathbf{Z})$ cancel out in the fraction, leaving a quantity that can be
computed analytically:
\begin{equation}
 \label{bound2}
\mathcal{F}_v = \int \mathcal{Q} \log 
	\dfrac{p(\dataMatrix|\mathbf{F}^Y) p(\mathbf{U}^Y) p(\latentMatrix|\mathbf{F}^X)
	p(\mathbf{U}^X) p(\mathbf{Z})}
	{\mathcal{Q}'} ,
\end{equation}
where 
$\mathcal{Q}'=q(\mathbf{U}^Y) q(\latentMatrix) q(\mathbf{U}^X) q(\mathbf{Z})$ 
and the above integration is with respect to 
$\{\latentMatrix, \mathbf{Z}, \mathbf{F}^Y, \mathbf{F}^X, \mathbf{U}^Y, \mathbf{U}^X\}$.
More specifically, we can break the logarithm in equation \eqref{bound2}
by grouping the variables of the fraction in such a way that the bound can be written as:
\begin{equation}
 \label{boundFinal}
 \mathcal{F}_v = \bfg_Y + \bfr_X + \mathcal{H}_{q(\latentMatrix)} - \text{KL}\left( q(\mathbf{Z}) \parallel p(\mathbf{Z}) \right)
\end{equation}
where $\mathcal{H}$ represents the entropy with respect to a distribution, $\text{KL}$ denotes the
Kullback -- Leibler divergence and, using $\la \cdot \ra$ to denote expectations,

\begin{align*}
 & \bfg_Y = g(\dataMatrix, \mathbf{F}^Y, \mathbf{U}^Y, \latentMatrix)  \nonumber \\
    & =  \la \log p(\dataMatrix|\mathbf{F}^Y) + \log \tfrac{p(\mathbf{U}^Y)}{q(\mathbf{U}^Y)} \ra_{p(\mathbf{F}^Y|\mathbf{U}^Y, \latentMatrix)q(\mathbf{U}^Y)q(\latentMatrix)} 
\end{align*}

\vspace{-14pt}

\begin{align}
 & \bfr_X   =  r(\latentMatrix, \mathbf{F}^X, \mathbf{U}^X, \mathbf{Z}) \nonumber \\
    & =  \la \log p(\latentMatrix|\mathbf{F}^X) + \log \tfrac{p(\mathbf{U}^X)}{q(\mathbf{U}^X)} \ra_{p(\mathbf{F}^X|\mathbf{U}^X,\mathbf{Z})q(\mathbf{U}^X)q(\latentMatrix)q(\mathbf{Z})}  \label{rx}
\end{align}


Both terms $\bfg_Y$ and $\bfr_X$ involve known Gaussian densities and are, thus, tractable.
The $\bfg_Y$ term is only associated with the leaves and, thus, is the same as the bound found for the Bayesian GP-LVM \citep{Titsias:bayesGPLVM10}. Since
it only involves expectations with respect to Gaussian distributions, the GP output variables are only involved in
a quantity of the form $\dataMatrix \dataMatrix^\T$. Further, as can be seen from the above equations, the function $r(\cdot)$
is similar to $g(\cdot)$ but it requires expectations with respect to densities of all of the variables involved (\ie with respect to all function inputs).
Therefore, $\bfr_X$ will involve $\latentMatrix$ (the outputs of the top layer) in a term
$
\la \latentMatrix \latentMatrix^\T\ra_{q(\latentMatrix)} = 
  \sum_{q=1}^Q \left[ \bfmu_q^X \left( \bfmu_q^X \right)^\T + \mathbf{S}_q^X \right] .
$


\section{Extending the hierarchy}
Although the main calculations were demonstrated in a simple
hierarchy, it is easy to extend the model vertically, \ie by adding
more hidden layers, or horizontally, \ie by considering conditional
independencies of the latent variables belonging to the same layer.
The first case only requires adding more $\bfr_X$ functions to the
variational bound, \ie instead of a single $\bfr_X$ term we will now
have the sum: $\sum_{h=1}^{H-1} \bfr_{X_h}$, where 
$\bfr_{X_h} = r(X_h, \mathbf{F}^{X_h}, \mathbf{U}^{X_h}, \latentMatrix_{h+1}), \;\; \latentMatrix_H = \mathbf{Z}$ .

\begin{figure}[ht]
\begin{center}
\subfigure[]{
\includegraphics[width=0.4\textwidth]{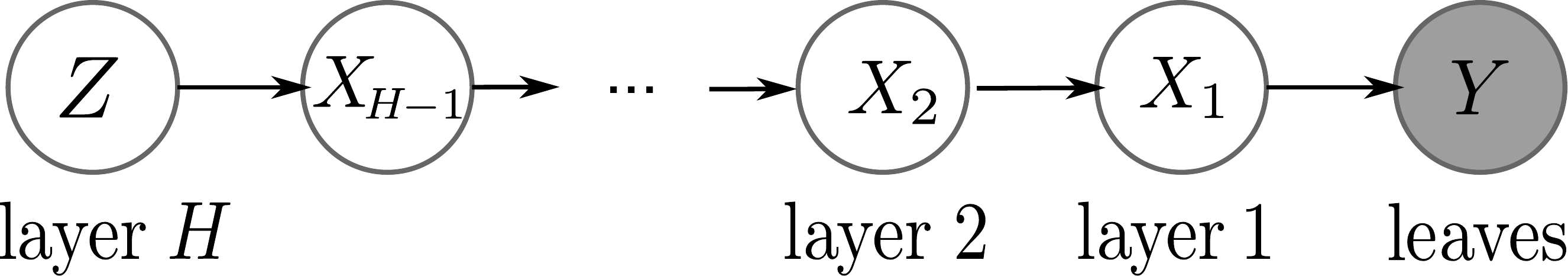}
	\label{hierarchy}
}
\newline
\subfigure[]{
	\includegraphics[width=0.18\textwidth]{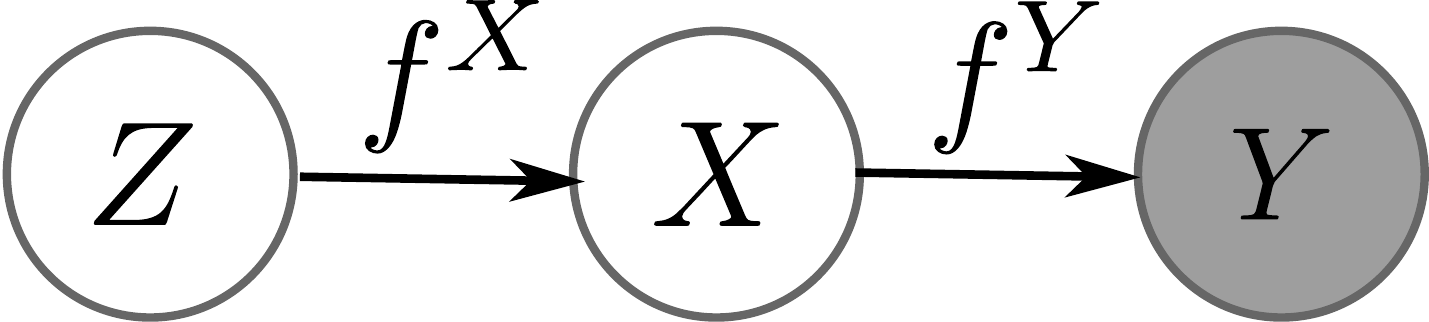}
	\label{hierarchySimple}
}
\subfigure[]{
	\includegraphics[width=0.42\textwidth]{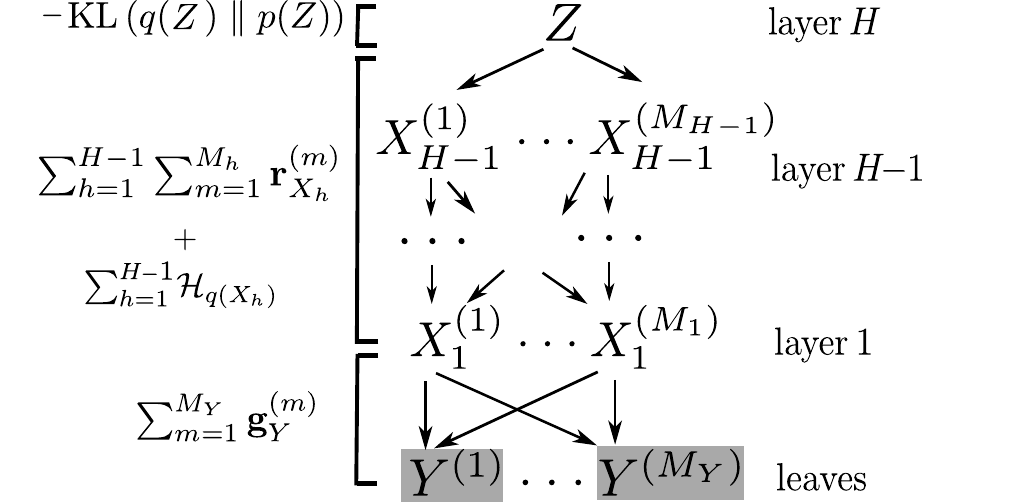}
	\label{hierarchyFull}
}
\end{center}
\vspace{-6pt}
\caption{
Different representations of the Deep GP model: 
\subref{hierarchy} shows the general architecture with a cascade
of $H$ hidden layers, \subref{hierarchySimple}
depicts a simplification of a two hidden layer hierarchy also demonstrating the
corresponding GP mappings and \subref{hierarchyFull} illustrates the most
general case where the leaves 
 and all intermediate nodes are allowed to
form conditionally independent groups. The terms of the objective \eqref{boundFull} corresponding
to each layer are included on the left.
}
\label{fig:grModel}
\vspace{-5pt}
\end{figure}

Now consider the horizontal expansion scenario and assume that we wish
to break the single latent space $\latentMatrix_h$, of layer $h$, to $M_h$
conditionally independent subsets.  As long as the variational
distribution $q(\latentMatrix_h)$ of equation \eqref{qX} is chosen to be factorised
in a consistent way, this is
feasible by just breaking the original $\bfr_{X_h}$ term of equation
\eqref{rx} into the sum $\sum_{m=1}^{M_h} \bfr_{X_h}^{(m)}$.  This
follows just from the fact that, due to the independence assumption,
it holds that $\log p(\latentMatrix_h|\latentMatrix_{h+1}) = \sum_{m=1}^{M_h} \log p(\latentMatrix_h^{(m)}
| \latentMatrix_{h+1})$.  Notice that the same principle can also be applied to
the leaves by breaking the $\bfg_Y$ term of the bound.  This scenario
arises when, for example we are presented with multiple different
output spaces which, however, we believe they have some
commonality. For example, when the observed data are coming from a
video and an audio recording of the same event.
Given the above, the variational bound for the most general version
of the model takes the form:
\begin{align}
\mathcal{F}_v & = 
    \sum_{m=1}^{M_Y} \bfg_Y^{(m)} +
    \sum_{h=1}^{H-1} \sum_{m=1}^{M_h} \bfr_{X_h}^{(m)} +
    \sum_{h=1}^{H-1} \mathcal{H}_{q(\latentMatrix_h)}  \nonumber \\
&-  \text{KL} \left( q(\mathbf{Z}) \parallel p(\mathbf{Z}) \right) .  \label{boundFull}
\end{align}
Figure \ref{fig:grModel}\subref{hierarchyFull} shows the association
of this objective function's terms with each layer of the hierarchy.
Recall that each $\bfr_{X_h}^{(m)}$ and $\bfg_Y^{(m)}$ term is associated with a
different GP and, thus, is coming with its own set of automatic
relevance determination (ARD) weights
(described in equation \eqref{rbfard}).

\subsection{Deep multiple-output Gaussian processes}

The particular way of extending the hierarchies horizontally, as
presented above, can be seen as a means of performing unsupervised
multiple-output GP learning.  This only requires assigning a different
$\bfg_Y$ term (and, thus, associated ARD weights) 
to each vector $\bfy_d$, where $d$ indexes the output dimensions. 
After training our model, we hope that the
columns of $\dataMatrix$ that encode similar information will be assigned
relevance weight vectors that are also similar.  This idea can be
extended to all levels of the hierarchy, thus obtaining a fully
factorised deep GP model.

This special case of our model makes the connection between our
model's structure and neural network architectures more obvious: the
ARD parameters play a role similar to the weights of neural networks,
while the latent variables play the role of neurons which learn
hierarchies of features.

\subsection{\label{complexity} Parameters and complexity}
In all graphical variants shown in figure \ref{fig:grModel}, every arrow
represents a generative procedure with a GP prior, corresponding to a set of parameters
$\{\tilde{\latentMatrix}, \boldsymbol \theta, \sigma_{\epsilon} \}$.
Each layer of latent variables corresponds to a variational distribution $q(\latentMatrix)$
which is associated with a set of variational means and covariances, as shown in equation \eqref{qX}.
The parent node can have the same form as equation \eqref{qX} or can be constrained with a more informative prior
which would couple the points of $q(\mathbf{Z})$.
For example, a dynamical prior would introduce $Q \times N^2$ parameters 
which can, nevertheless, be reparametrized using less variables \citep{Damianou:vgpds11}.
However, as is evident from equations \eqref{varDistr} and \eqref{bound2},
the inducing points and the parameters of $q(\latentMatrix)$ and $q(\mathbf{Z})$ are \textit{variational}
rather than model parameters, something which significantly helps in regularizing the problem.
Therefore, adding more layers to the hierarchy does not introduce many more model parameters.
Moreover, as in common sparse methods for Gaussian processes \citep{Titsias09}, the complexity of 
each generative GP mapping is reduced from the typical $O(N^3)$ to $O(N M^2)$.


\section{Demonstration}

In this section we demonstrate the deep GP model in toy and real-world
data sets. For all experiments, the model is initialised by performing
dimensionality reduction in the observations to obtain the first
hidden layer and then repeating this process greedily for the next
layers. To obtain the stacked initial spaces we experimented with PCA
and the Bayesian GP-LVM, but the end result did not vary
significantly. Note that the usual process in deep learning is to seek
a dimensional expansion, particularly in the lower layers. In deep GP
models, such an expansion \emph{does} occur between the latent layers
because there is an infinite basis layer associated with the
GP between each latent layer.

\subsection{Toy Data}

\begin{figure}[ht]
\begin{center}
\subfigure[]{\includegraphics[width=0.1\textwidth]{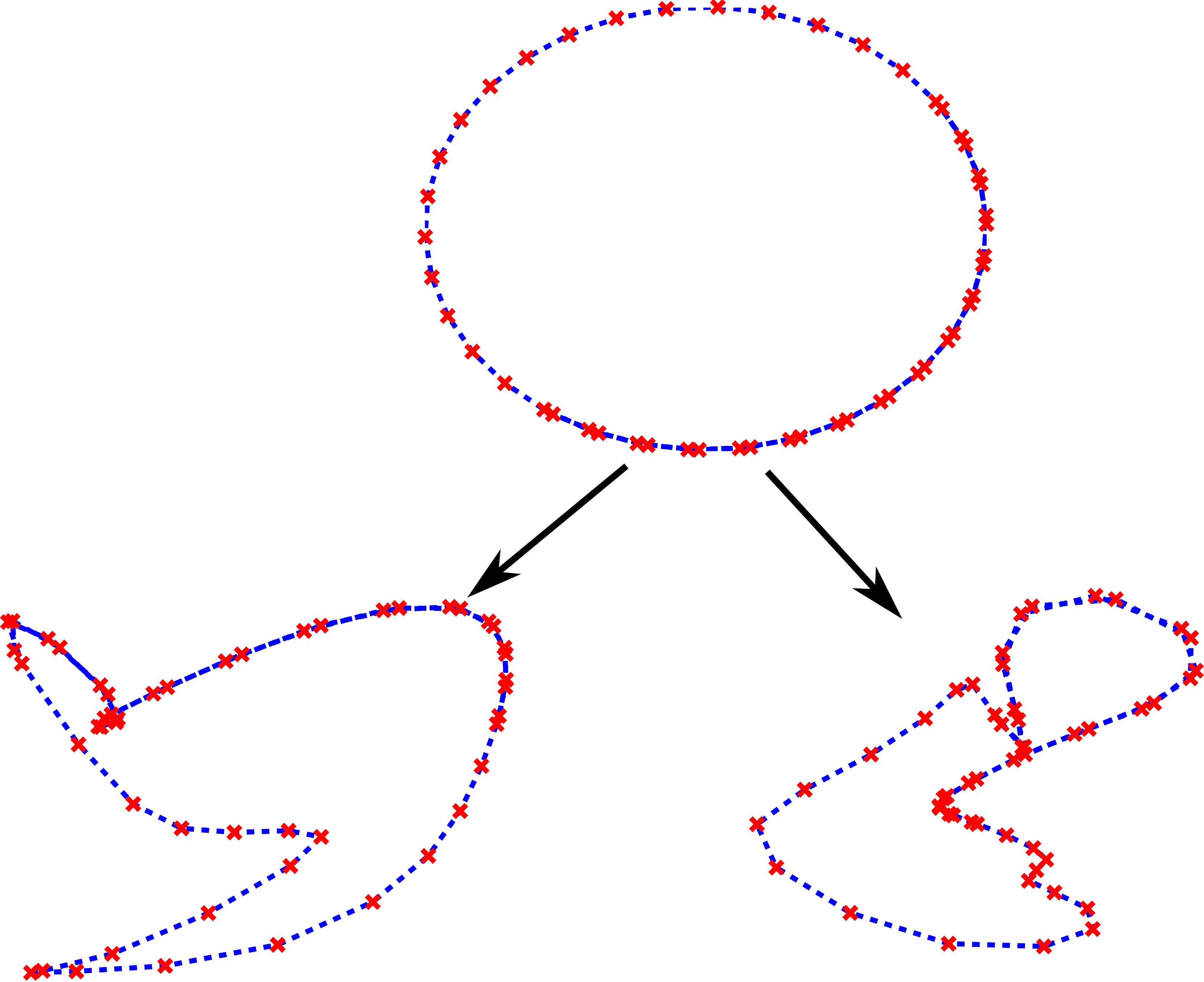}} \hspace{5pt}
\subfigure[]{\includegraphics[width=0.1\textwidth]{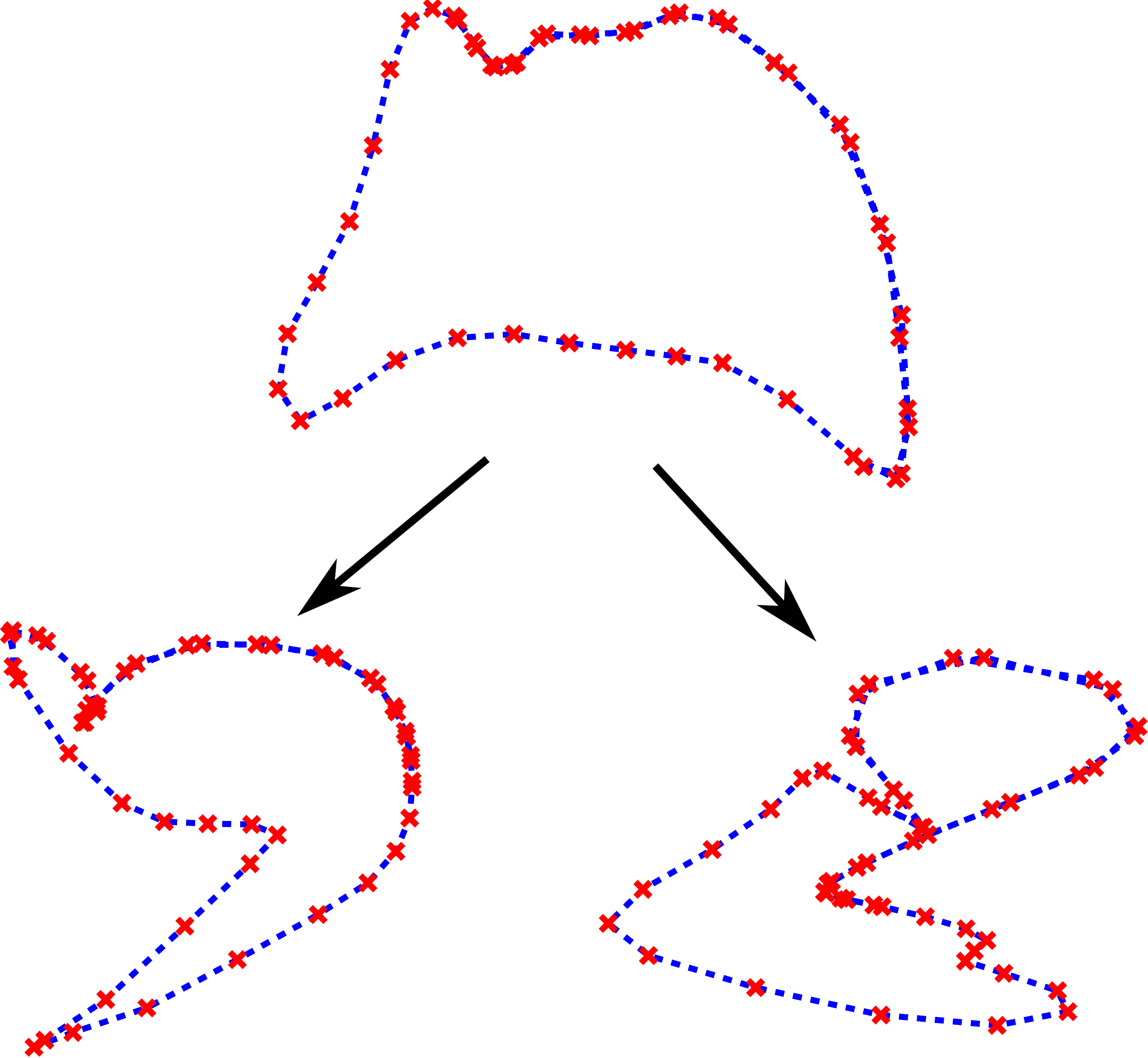} }\hspace{5pt}
\subfigure[]{\includegraphics[width=0.1\textwidth]{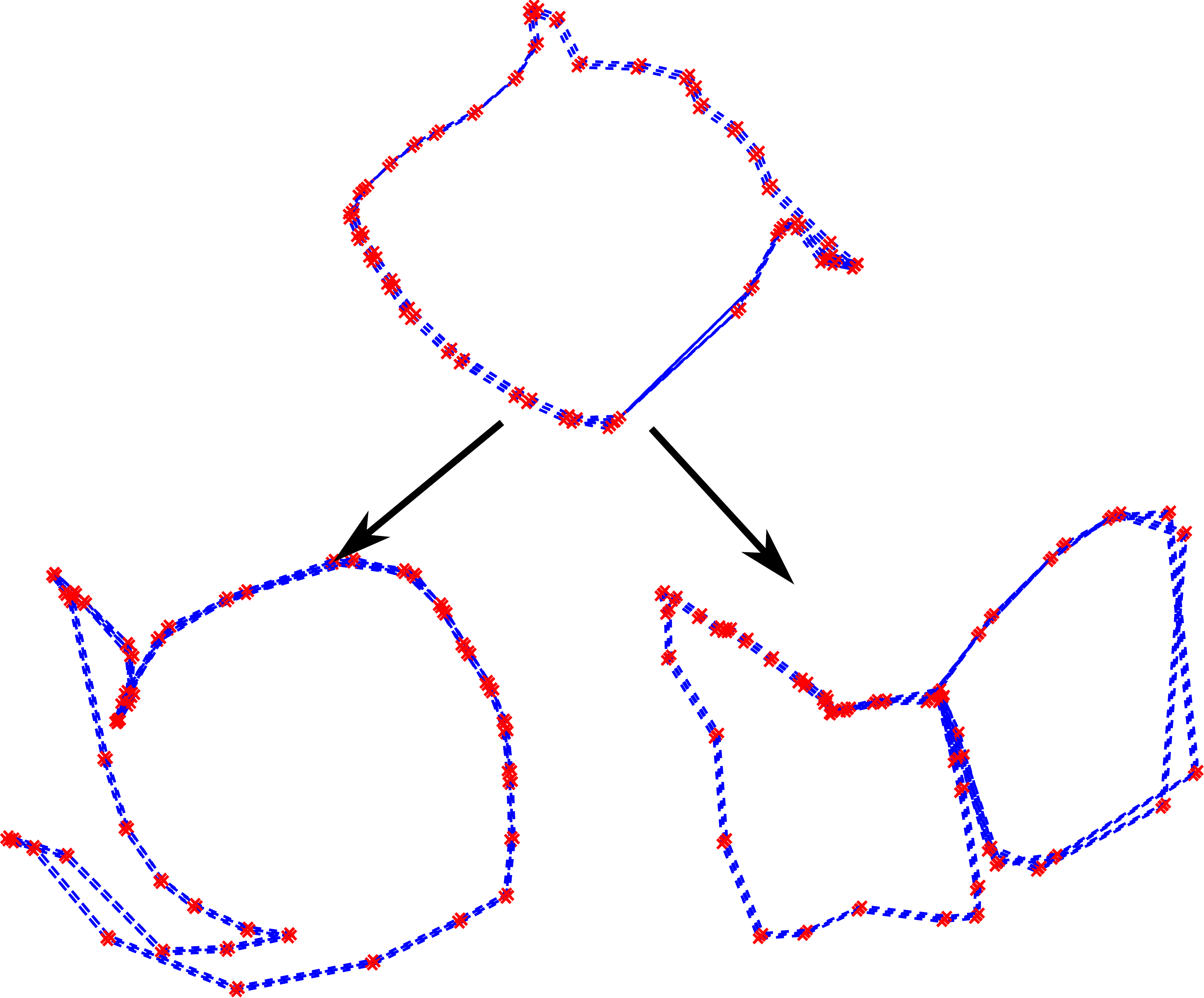} }\hspace{5pt}
\subfigure[]{\includegraphics[width=0.11\textwidth]{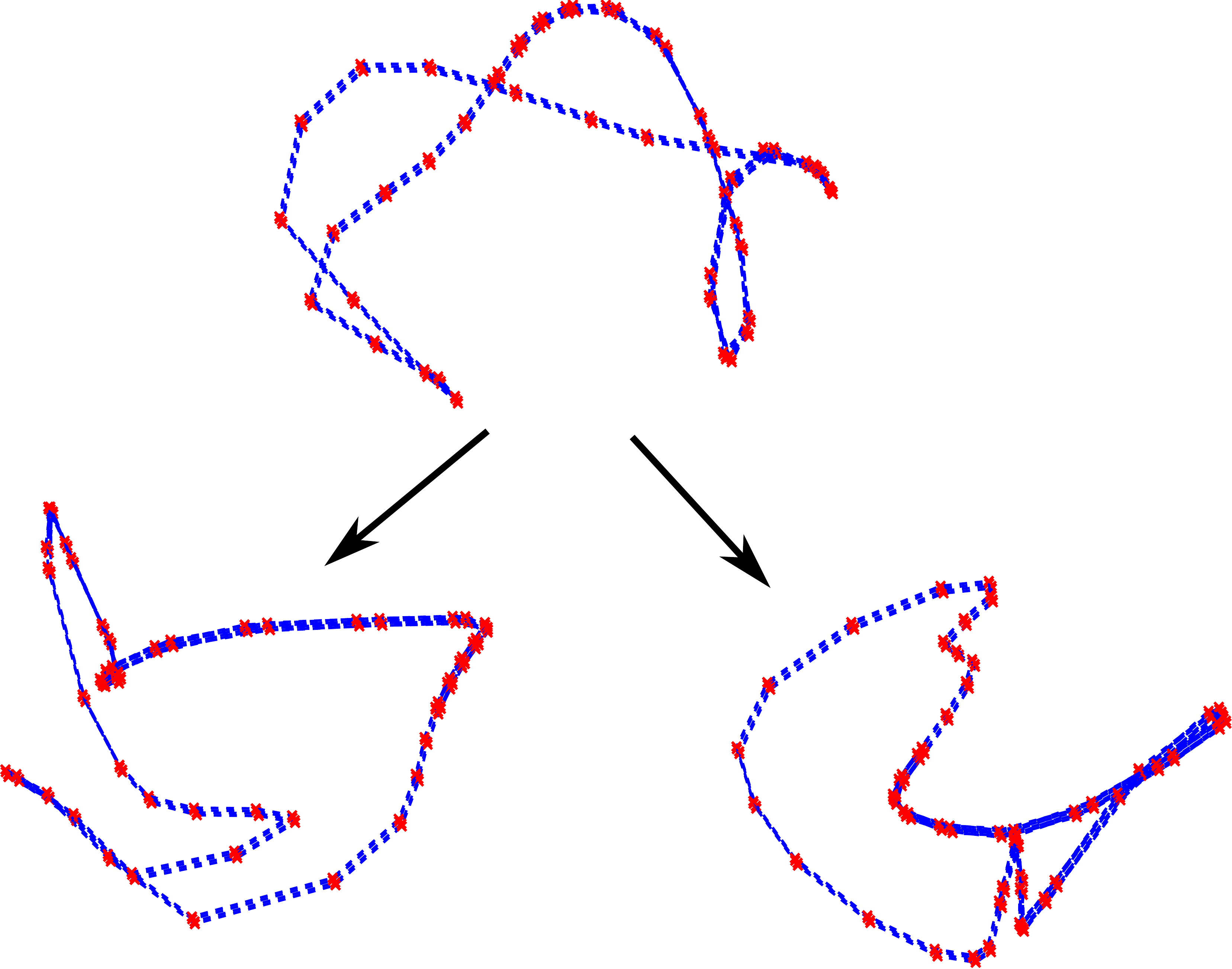} }
\end{center}
\vspace{-10pt}
\caption{ Attempts to reconstruct the real data (fig. (a)) with our
  model (b), stacked Isomap (c) and stacked PCA (d). Our model can
  also find the correct dimensionalities automatically.  }
\label{fig:toyData}
\vspace{-2pt}
\end{figure}

We first test our model on toy data, created by sampling from a
three-level stack of GPs. Figure \ref{fig:toyData} (a) depicts the
true hierarchy: 
from the top latent layer two intermediate latent signals are
generated. These, in turn, together generate $10$-dimensional
observations (not depicted) through sampling of another GP. These
observations are then used to train the following models: a deep GP, a
simple stacked Isomap  \citep{Tenenbaum:isomap00} 
and a stacked PCA method, the results of
which are shown in figures \ref{fig:toyData} (b, c, d)
respectively. From these models, only the deep GP marginalises the
latent spaces and, in contrast to the other two, it is not given any
information about the dimensionality of each true signal in the
hierarchy; instead, this is learnt automatically through ARD.  As can
be seen in figure \ref{fig:toyData}, the deep GP finds the
correct dimensionality for each hidden layer, but it also discovers
latent signals which are closer to the real ones. 
This result is encouraging, as it indicates that 
the model can recover the ground truth when samples from it are taken, 
%
 and gives
 confidence in the variational 
learning procedure.

We next tested our model on a toy regression problem. A deep
regression problem is similar to the unsupervised learning problem
we have described, but in the uppermost layer we make observations of
some set of inputs. For this simple example we created a toy data set by
stacking two Gaussian processes as follows: the first Gaussian process
employed a covariance function which was the sum of a linear and an
quadratic exponential kernel and received as input an equally spaced
vector of $120$ points. We generated $1$-dimensional samples from the
first GP and used them as input for the second GP, which employed a
quadratic exponential kernel. Finally, we generated $10$-dimensional
samples with the second GP, thus overall simulating a warped
process. The final data set was created by simply ignoring the
intermediate layer (the samples from the first GP) and presenting to
the tested methods only the continuous equally spaced input given to
the first GP and the output of the second GP. To make the data set more
challenging, we randomly selected only $25$ datapoints for the training set and
left the rest for the test set.

\begin{figure}[ht]
\begin{center}
\subfigure[]{\includegraphics[width=0.16\textwidth]{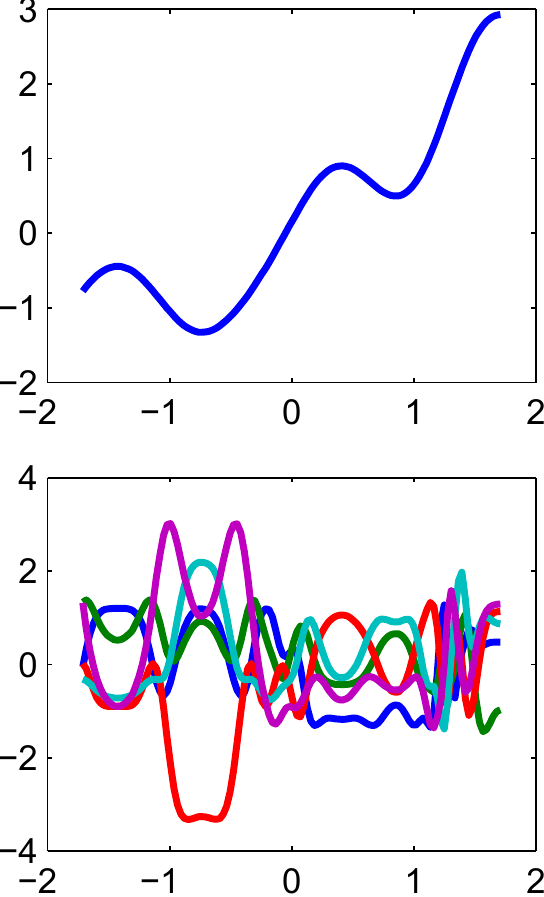}\label{toyRegData}} \hspace{5pt}
\subfigure[]{\includegraphics[width=0.3\textwidth]{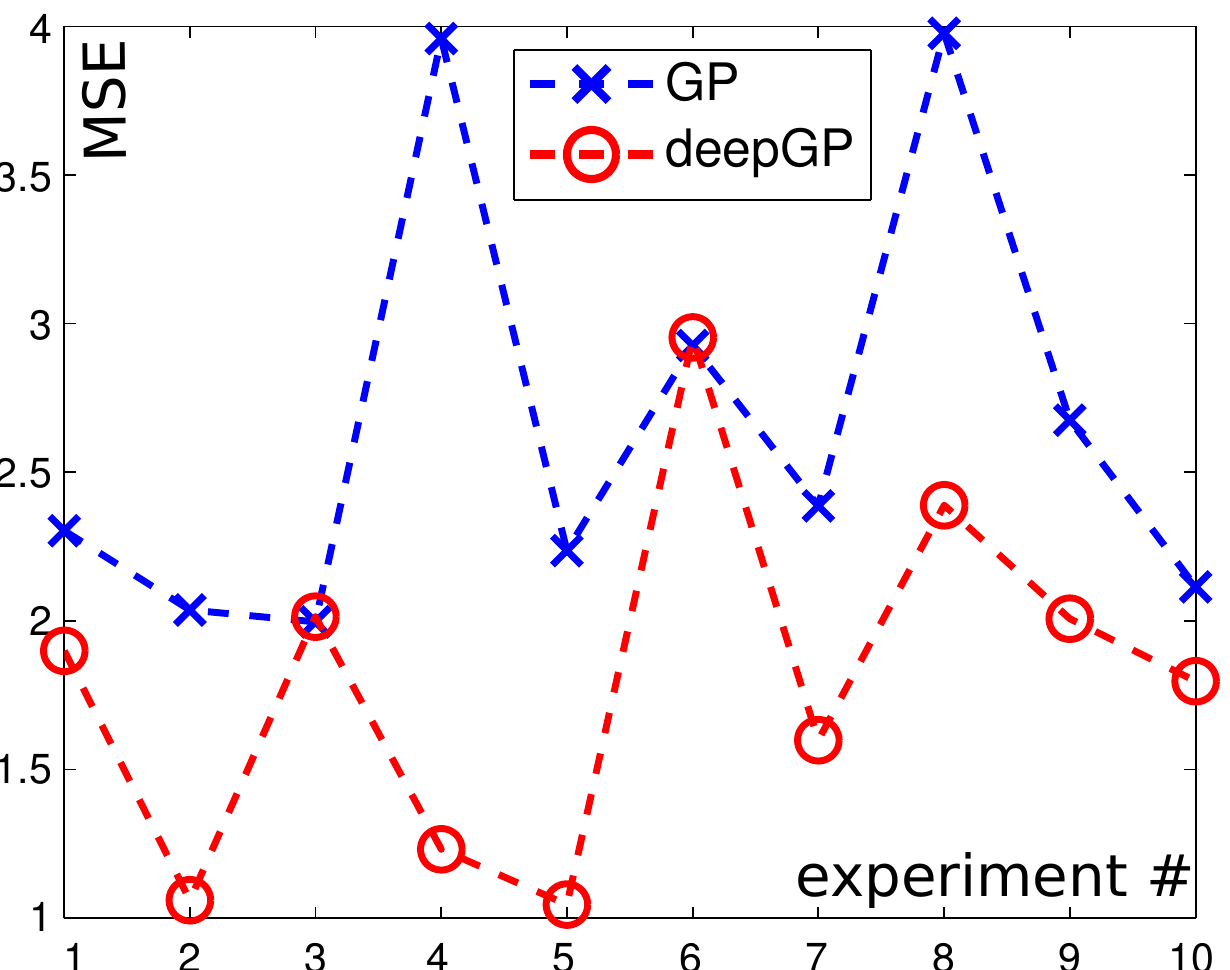}\label{toyRegResults}}
\end{center}
\vspace{-10pt}
\caption{\subref{toyRegData} shows the toy data created for the regression experiment. The top plot
shows the (hidden) warping function and bottom plot shows the final (observed) output.
\subref{toyRegResults} shows the results obtained over each experiment repetition.
}
\label{fig:toyDataRegression}
\vspace{-4pt}
\end{figure}

Figure \ref{fig:toyDataRegression} nicely illustrates the effects of
sampling through two GP models, nonstationarity and long range
correlations across the input space become prevalent. A data set of
this form would be challenging for traditional approaches because of
these long range correlations. Another way of thinking of data like this is as a 
nonlinear warping of the input space to the GP. Because this type of deep GP only
contains one hidden layer, it is identical to the model developed by 
\citep{Damianou:vgpds11} (where the input given at the top layer of their model was a time vector, but
their code is trivially generalized). The additional contribution in this paper 
will be to provide a more complex deep hierarchy, but still learn the underlying 
representation correctly. To this end we applied a standard GP (1 layer less than the
actual process that generated the data) and a deep GP with two hidden
layers (1 layer more than the actual generating process). We repeated
our experiment 10 times, each time obtaining different samples from
the simulated warped process and different random training splits.  Our results
show that the deep GP predicted better the unseen data, as can be seen
in figure \ref{fig:toyDataRegression}\subref{toyRegResults}.
The results, therefore, suggest that our deep model can at the same
time be flexible enough to model difficult data as well as robust,
when modelling data that is less complex than that representable by the hierarchy.
We assign these characteristics to the Bayesian learning approach that deals with 
capacity control automatically.

\subsection{Modeling human motion}

\begin{figure}[ht]
\begin{center}
\subfigure[]{ \includegraphics[width=0.11\textwidth]{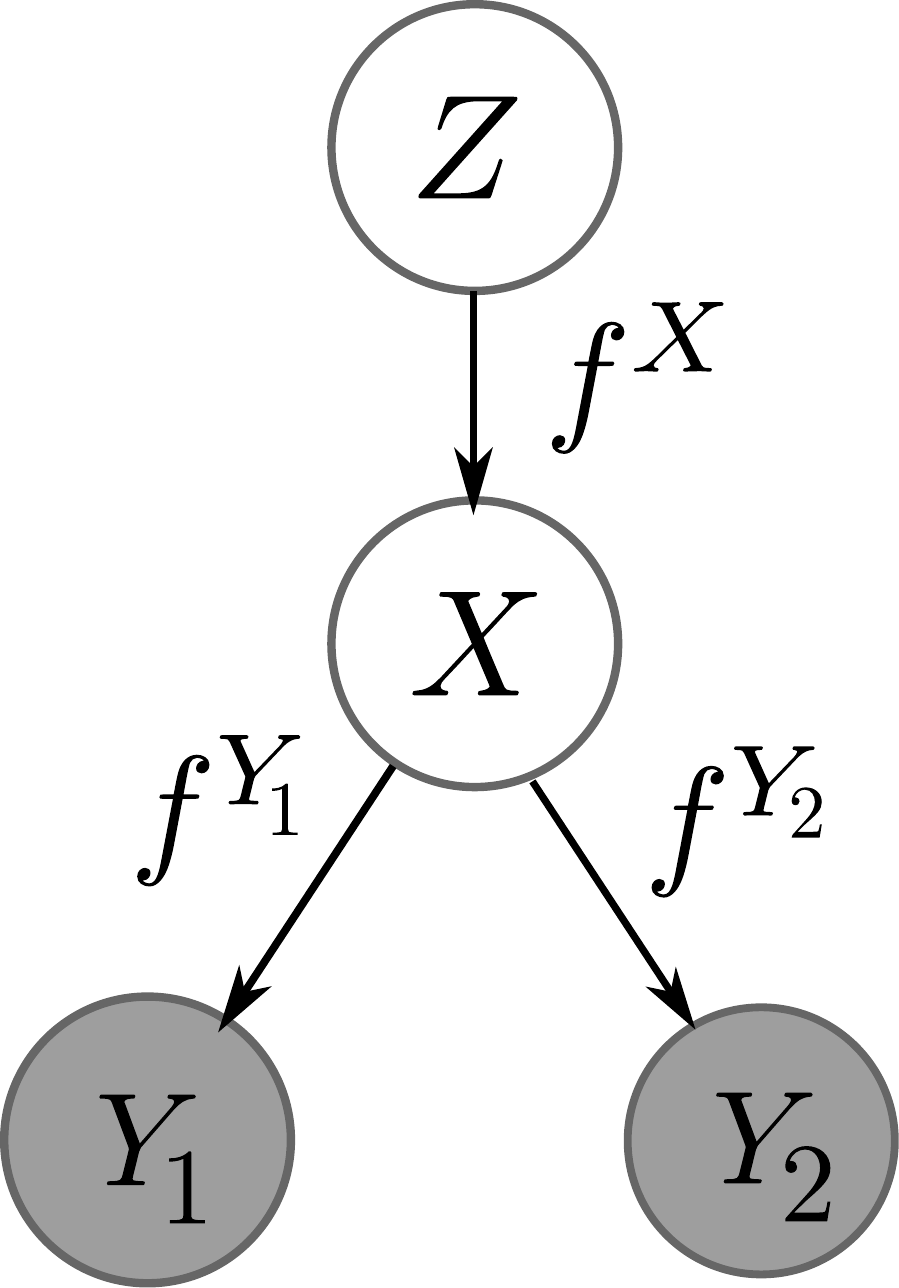} \label{hierHighFive}}
	\hspace{8pt}
\subfigure[]{ \includegraphics[width=0.15\textwidth]{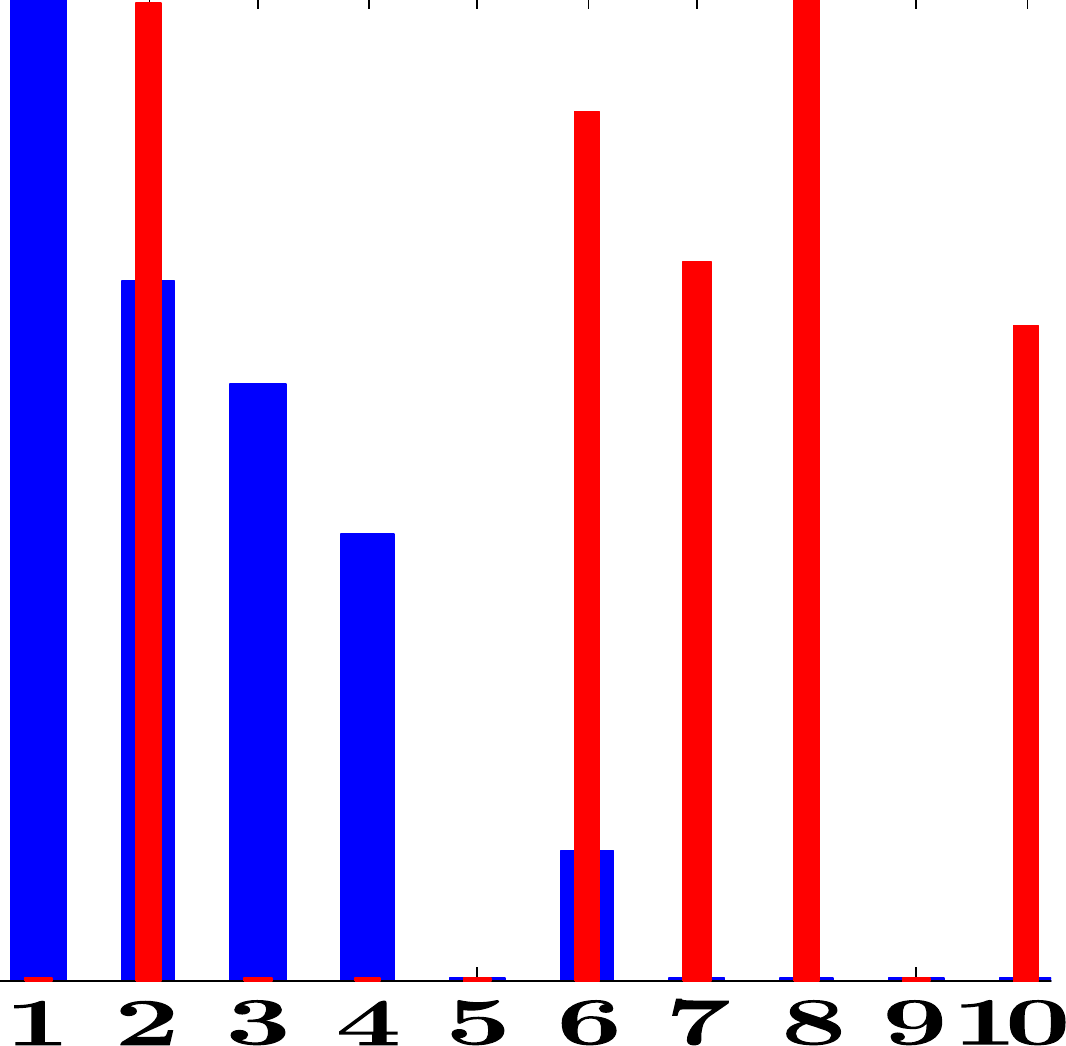} \label{highFiveScales1}}
\hspace{2pt}
\subfigure[]{ \includegraphics[width=0.15\textwidth]{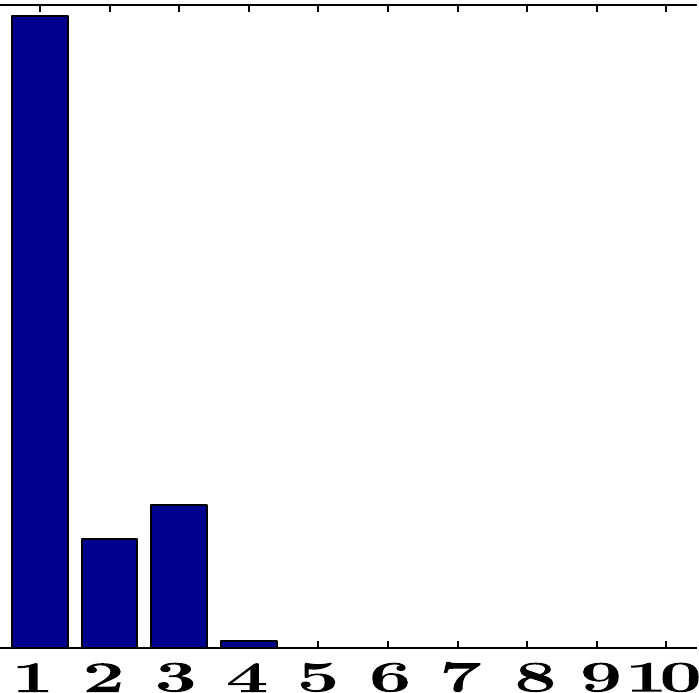} \label{highFiveScales2}}
\end{center}
\vspace{-6pt}
\caption{
Figure \subref{hierHighFive} shows the deep GP model employed.
Figure \subref{highFiveScales1} shows the ARD weights for $f^{Y_1}$ (blue/wider bins) and $f^{Y_2}$ (red/thinner bins)
and figure \subref{highFiveScales2} those for $f^X$. 
}
\label{fig:highFiveScales}
\end{figure}

\begin{figure}[ht]
 \begin{center}
 \subfigure[]{ \includegraphics[width=0.06\textwidth]{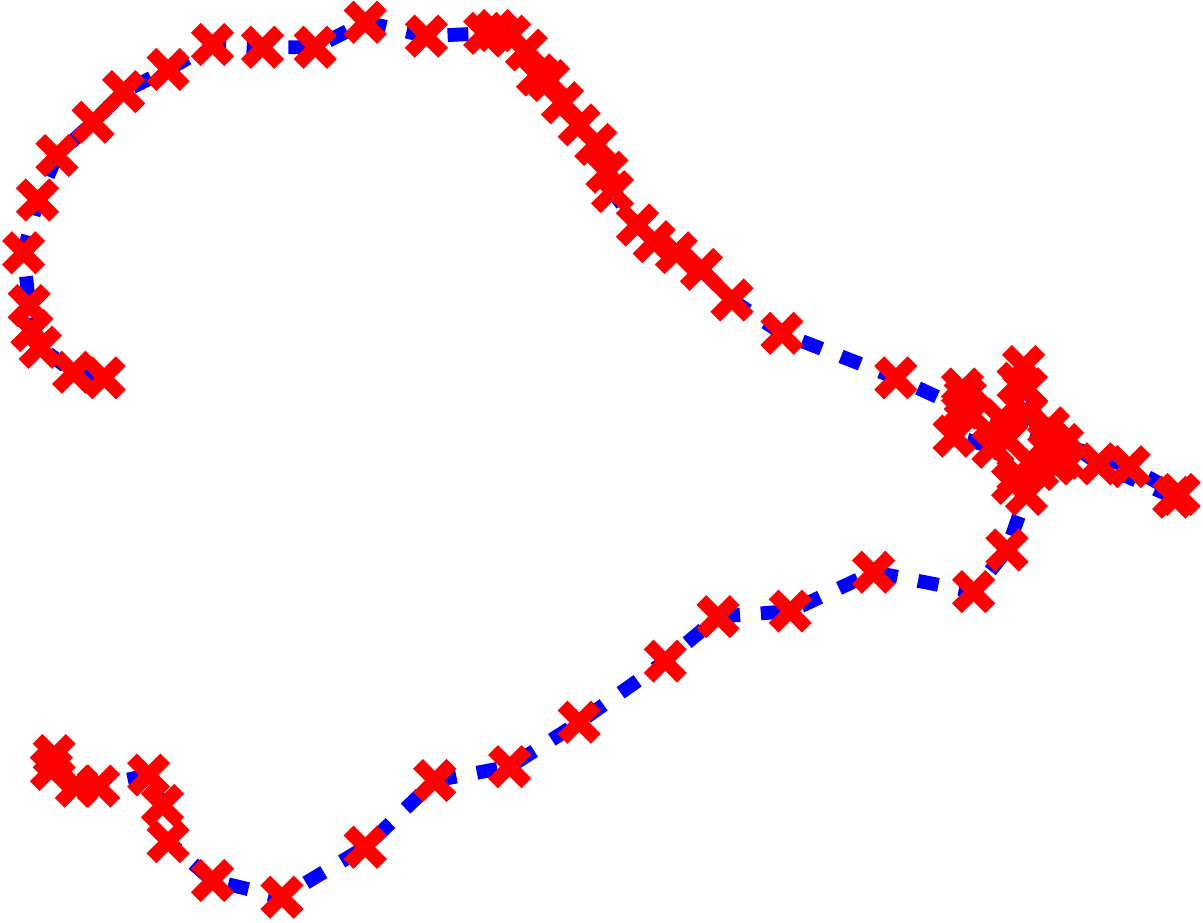} }
 \subfigure[]{ \includegraphics[width=0.06\textwidth]{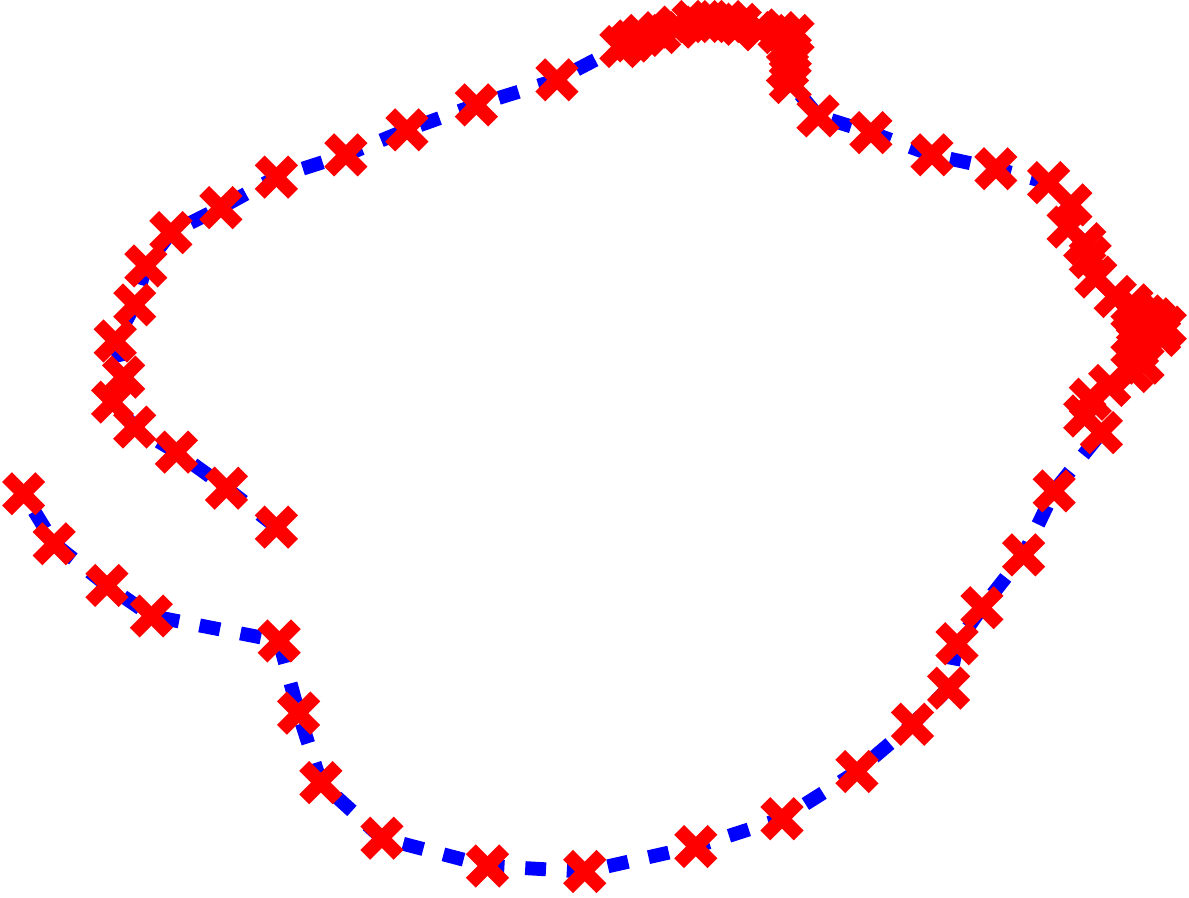} } 
 \subfigure[]{ \includegraphics[width=0.06\textwidth]{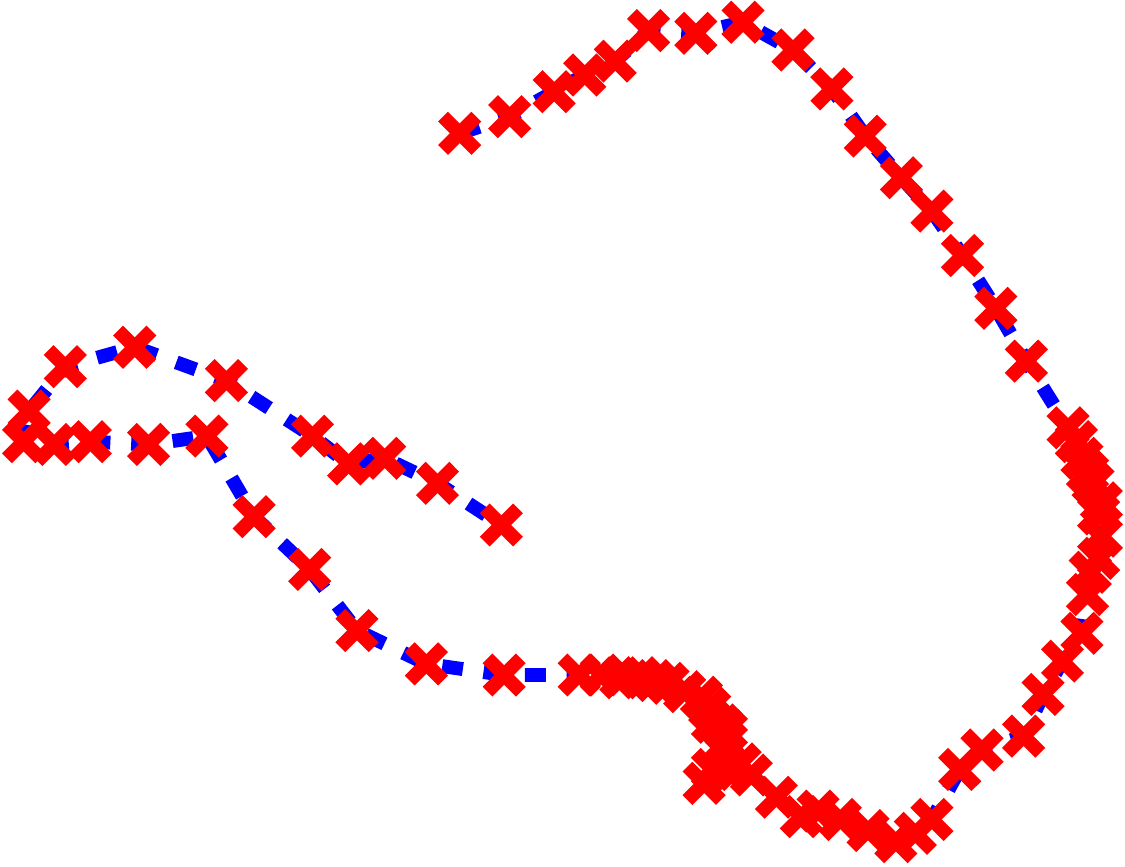} }
 \subfigure[]{ \includegraphics[width=0.06\textwidth]{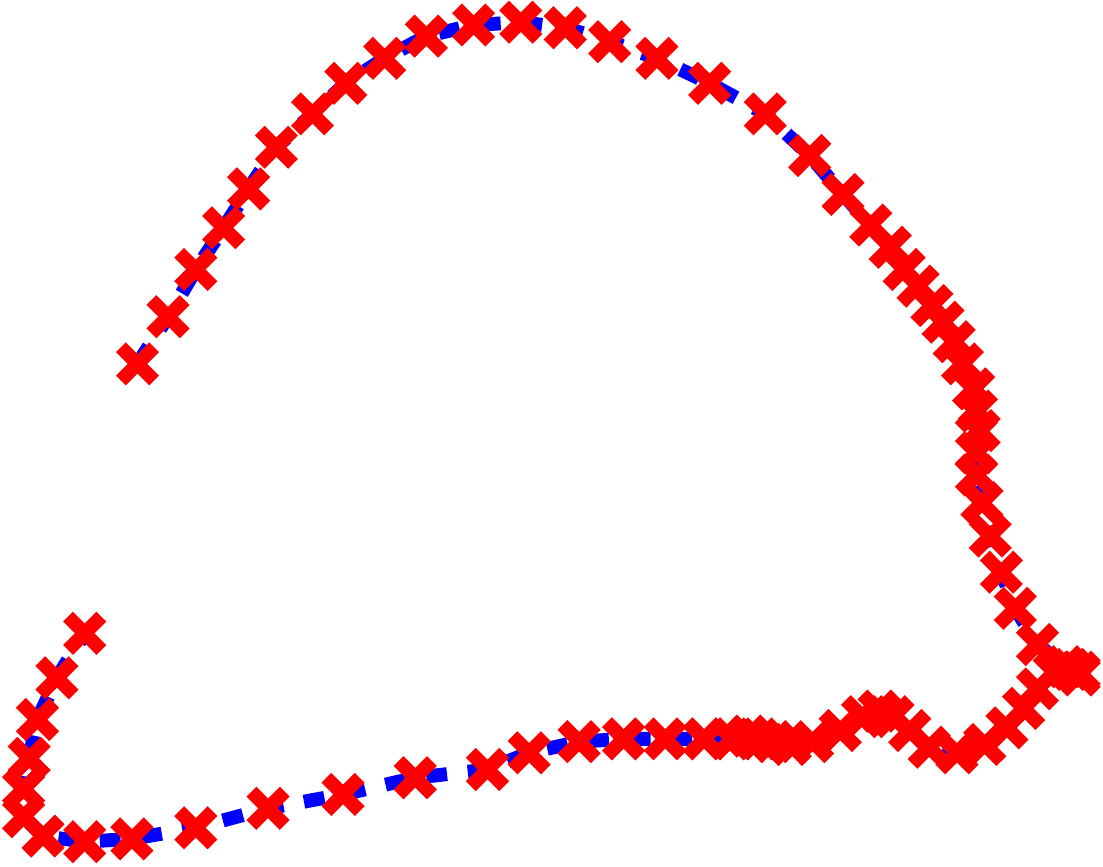} } 
 \subfigure[]{ \includegraphics[width=0.06\textwidth]{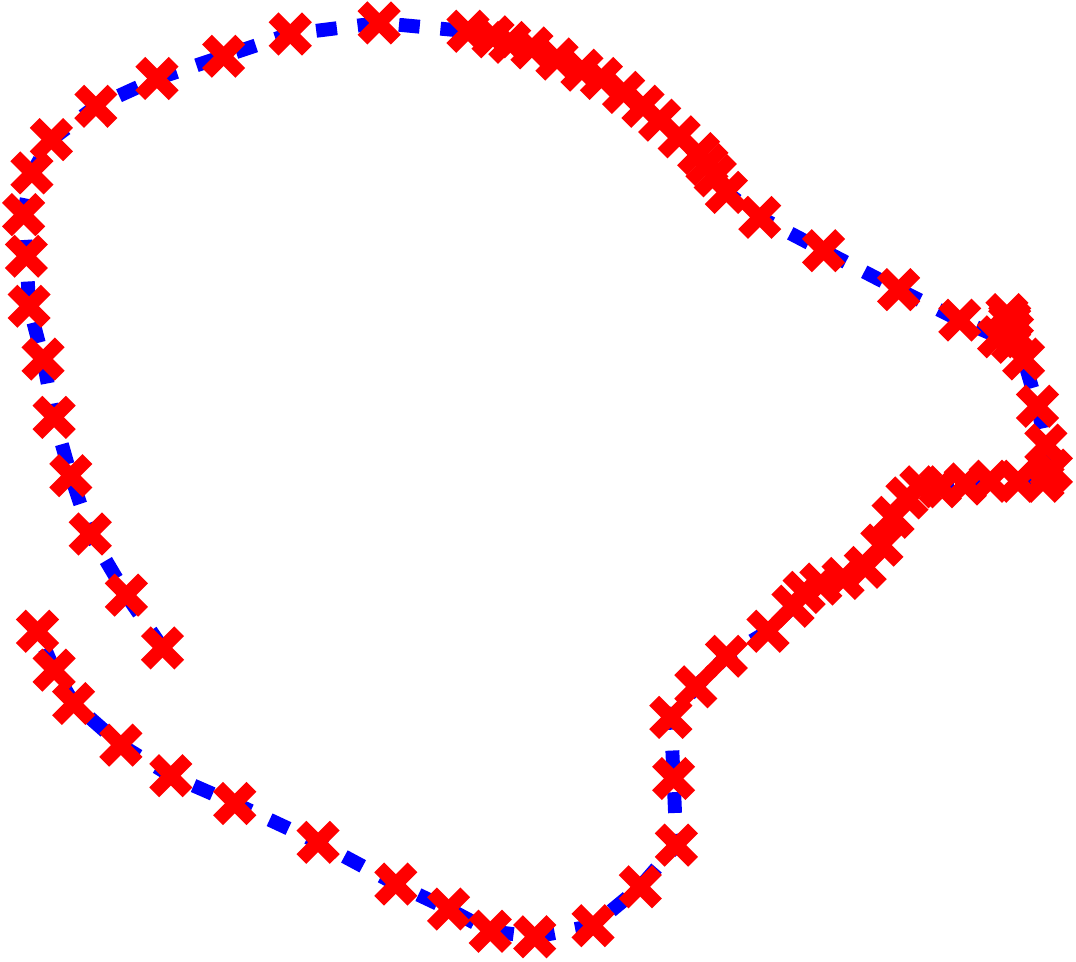} } 
 \subfigure[]{ \includegraphics[width=0.06\textwidth]{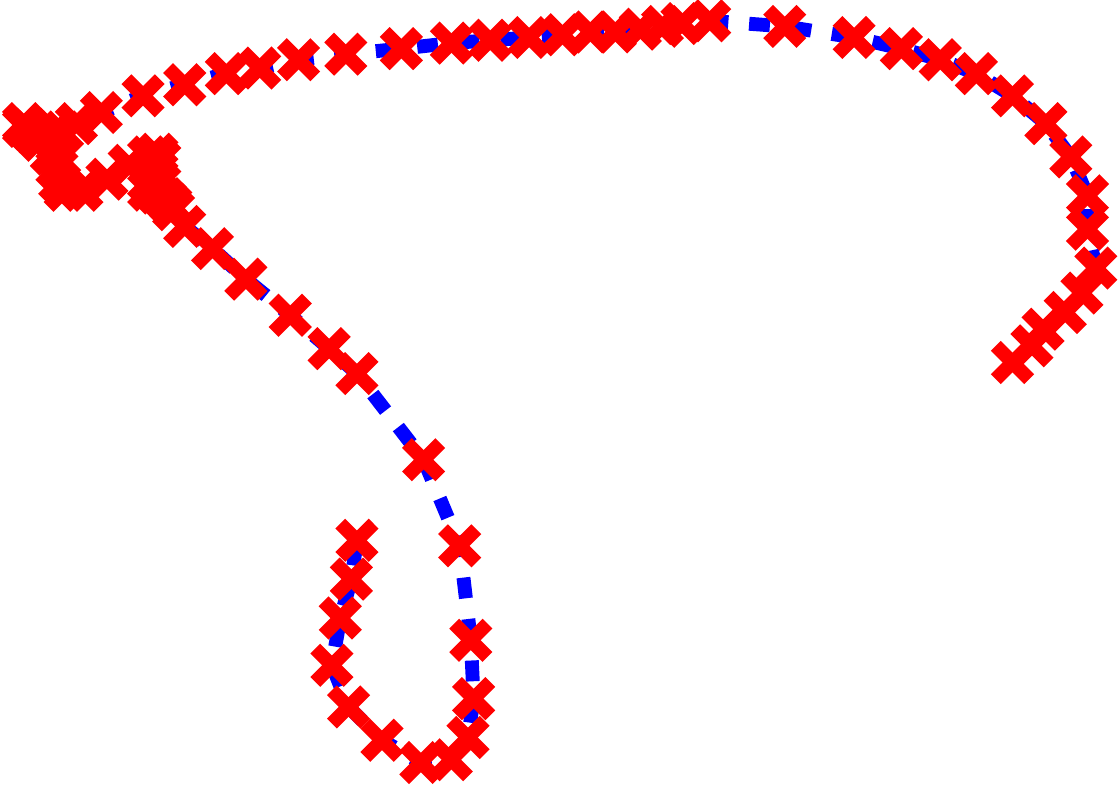} } 
 \end{center}
  \vspace{-6pt}
\caption{
Left (a,b,c): projections of the latent spaces discovered by our model, Right (d,e,f): the full latent space learned for 
the model of \cite{Lawrence:hgplvm07}.
}
\label{fig:demHighFiveSpaces}
 \end{figure}

For our first demonstration on real data we recreate a motion capture
data experiment from \cite{Lawrence:hgplvm07}. They used data from the
CMU MOCAP database 
representing two
subjects walking towards each other and performing a `high-five'. The
data contains 78 frames of motion and each character has 62
dimensions, leading to 124 dimensions in total (i.e. more dimensions than
data). To account for the correlated motions of the subjects we
applied our method with a two-level hierarchy where the two
observation sets were taken to be conditionally independent given
their parent latent layer. In the layer closest to the data we associated 
each GP-LVM with a different set of ARD parameters, allowing the layer 
above to be used in different ways for each character. In this approach 
we are inspired by the shared GP-LVM structure of \cite{Damianou:manifold12}
which is designed to model loosely correlated data sets within the same model. 
The end result was that we obtained three optimised sets
of ARD parameters: one for each modality of the bottom layer 
(fig. \ref{fig:highFiveScales}\subref{highFiveScales1}),
and one 
for the top node 
(fig.
\ref{fig:highFiveScales}\subref{highFiveScales2}). 
Our model discovered
a common subspace in the intermediate layer, since for dimensions $2$
and $6$ both ARD sets have a non-zero value. This is expected, as the
two subjects perform very similar motions with opposite
directions. The ARD weights are also a means of automatically
selecting the dimensionality of each layer and subspace. This kind of
modelling is impossible for a MAP method like
\citep{Lawrence:hgplvm07} which requires the exact latent structure to
be given a priori.  The full latent space learned by the
aforementioned MAP method is plotted in figure
\ref{fig:demHighFiveSpaces} (d,e,f), where fig. (d) corresponds to the
top latent space and each of the other two encodes information for
each of the two interacting subjects. Our method is not constrained to
two dimensional spaces, so for comparison we plot two-dimensional
projections of the dominant dimensions of each subspace in figure
\ref{fig:demHighFiveSpaces} (a,b,c). The similarity of the latent
spaces is obvious.  In contrast to \cite{Lawrence:hgplvm07}, we did
not have to constrain the latent space with dynamics in order to
obtain results of good quality.

Further, we can sample from these spaces to see what kind of information they encode. Indeed, we observed that the top layer
generates outputs which correspond to different variations of the whole sequence, while when sampling from the
first layer we obtain outputs which only differ in a small subset of the output dimensions, \eg those corresponding
to the subject's hand.

%
%

\subsection{Deep learning of digit images}

Our final experiment demonstrates the ability of our model to learn
latent features of increasing abstraction and we demonstrate the
usefulness of an analytic bound on the model evidence as a means of
evaluating the quality of the model fit for different choices of the
overall depth of the hierarchy. Many deep learning approaches are
applied to large digit data sets such as MNIST. Our specific intention
is to explore the utility of deep hierarchies when the digit data set
is \emph{small}. We subsampled a data set consisting of $50$ examples
for each of the digits $\{0,1,6\}$ taken from the USPS handwritten
digit database. Each digit is represented as an image in $16 \times
16$ pixels. We experimented with deep GP models of depth ranging from
$1$ (equivalent to Bayesian GP-LVM) to $5$ hidden layers and evaluated
each model by measuring the nearest neighbour error in the latent
features discovered in each hierarchy. We found that the lower bound
on the model evidence increased with the number of layers as did
the quality of the model in terms of nearest neighbour errors
\footnote{As parameters increase linearly in the deep GP with latent units,
we also considered the Bayesian Information Criterion, but we found that it had no effect on
the ranking of model quality.}.  Indeed, the single-layer model
made $5$ mistakes even though it automatically decided to use $10$
latent dimensions
and the quality of the trained models was increasing with the number of hidden layers.
Finally, only one point had a
nearest neighbour of a different class in the $4-$dimensional top
level's feature space of a model with depth $5$.  
A $2D$ projection of
this space is plotted in fig.7. 
 The ARD
weights for this model are depicted in fig. 6. 

Our final goal is to demonstrate that, as we rise in the hierarchy, features of 
increasing abstraction are accounted for. To this end, we
generated outputs by sampling from each hidden layer. The samples are shown in 
figure 8. 
There, it can be seen that the lower levels encode local features 
whereas the higher ones encode more
abstract information.

\begin{figure}[ht]
\begin{center}
\includegraphics[width=0.47\textwidth]{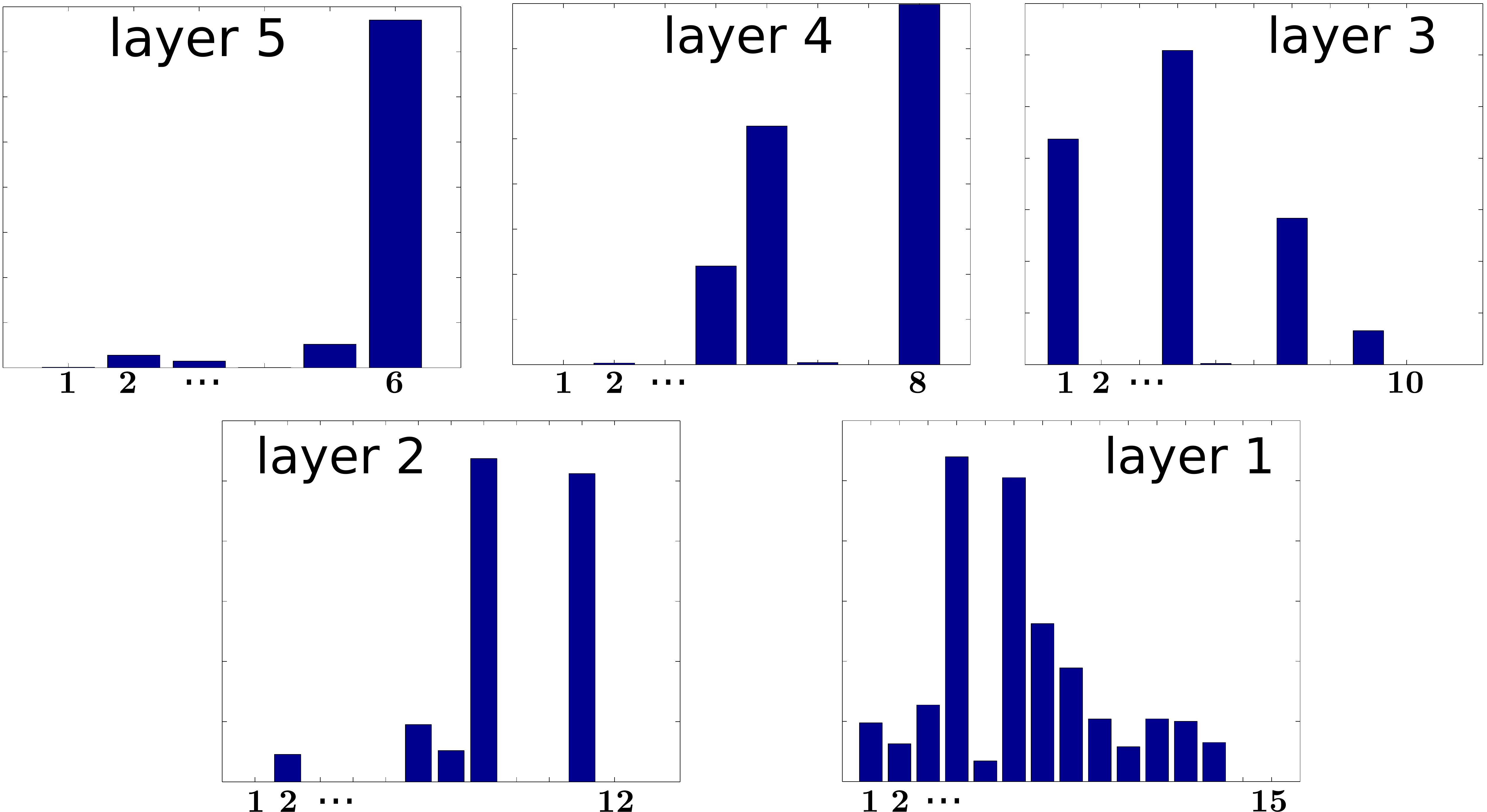}
\end{center}
\vspace{-4pt}
\caption{
The ARD weights of a deep GP with $5$ hidden layers as learned for the digits experiment.
}
\label{fig:uspsScales}
\vspace{-1pt}
\end{figure}

\begin{figure}[ht]
\begin{center}
 \includegraphics[width=0.39\textwidth]{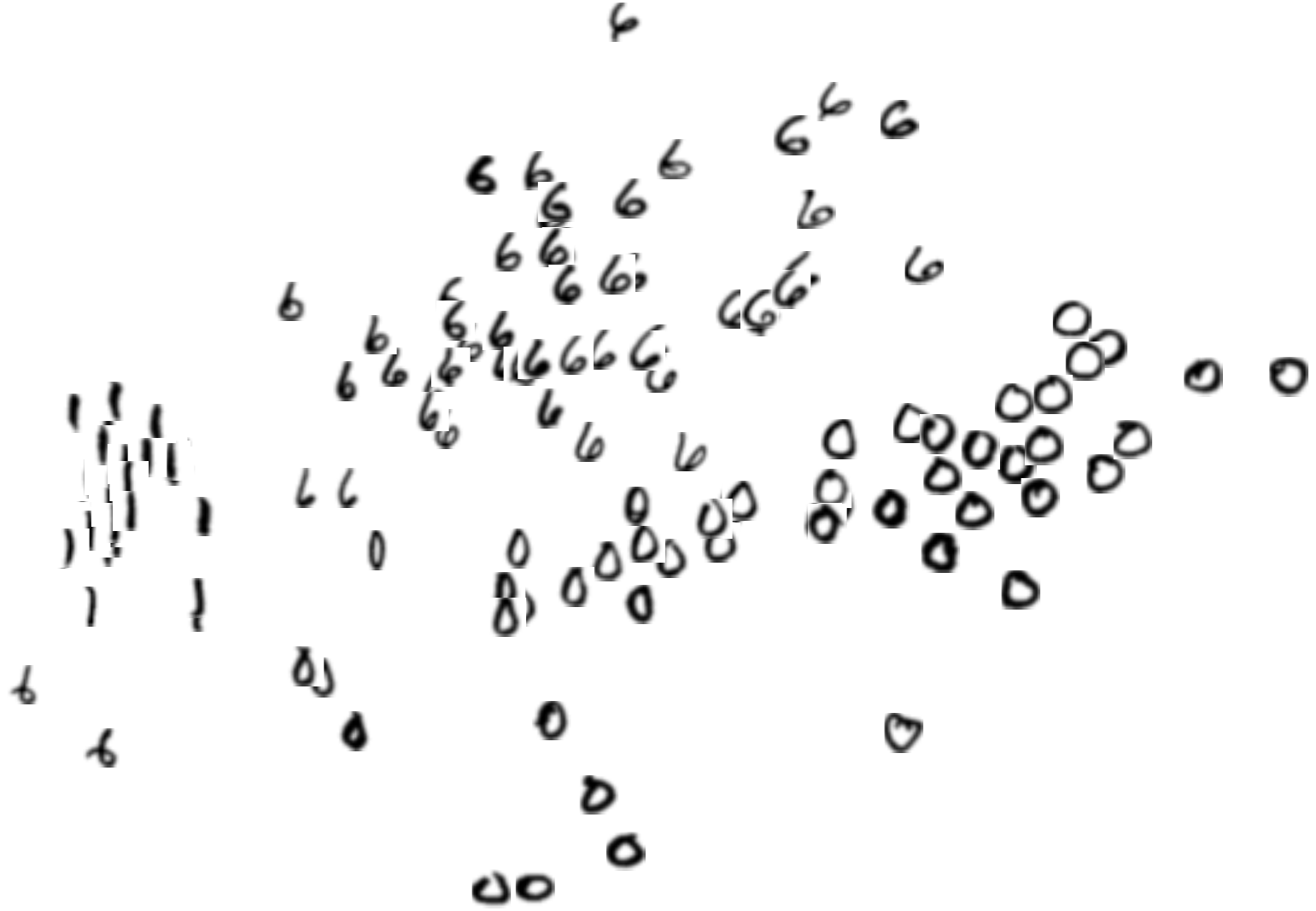} \label{fig:digitsSpace} 
\end{center}
\vspace{-6pt}
\caption{
The nearest neighbour class separation test on a deep GP model with depth $5$. 
}
\vspace{-1pt}
\end{figure}

\begin{figure}[ht]
\begin{center}
\includegraphics[width=0.42\textwidth]{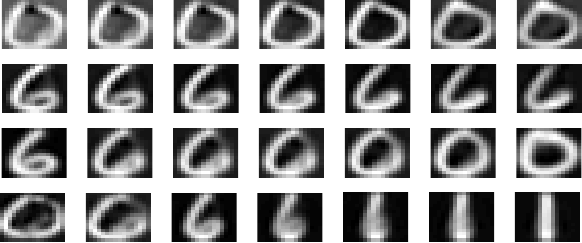} \label{fig:digits_samples} 
\end{center}
\vspace{-5pt}
\caption{
The first two rows (top-down) show outputs obtained when sampling from layers
$1$ and $2$ respectively and encode very local features, \eg explaining if a ``$0$'' has a closed circle
or how big the circle of a ``$6$'' is. We found many more local features when we sampled from different dimensions. Conversely,
when we sampled from the two dominant dimensions of the parent node (two rows in the bottom) we got much more varying
outputs, \ie the higher levels indeed encode much more abstract information.
}
\end{figure}


\section{Discussion and future work}

We have introduced a framework for efficient Bayesian training of
hierarchical Gaussian process mappings. 
Our
approach approximately marginalises out the latent space, thus
allowing for automatic structure discovery in the hierarchy. The
method was able to successfully learn a hierarchy of features which
describe natural human motion and the pixels of handwritten
digits. Our variational lower bound selected a deep hierarchical
representation for handwritten digits even though the data in our
experiment was relatively scarce (150 data points).
%
We gave persuasive evidence that deep GP models are powerful
enough to encode abstract information even for smaller data
sets. Further exploration 
could include testing
the model on other inference tasks, such as class conditional density
estimation to further validate the ideas. Our method can also be used
to improve existing deep algorithms, something which we plan to
further investigate by incorporating ideas from past
approaches. 
Indeed, previous efforts to combine GPs with deep structures were
successful at unsupervised pre-training
\citep{Erhan:unsupervisedPreTraining} or guiding
\citep{nonParametricGuidanceForAutoencoders} of traditional deep
models.

Although the experiments presented here considered only up to 5
layers in the hierarchy, the methodology is directly applicable to
deeper architectures, with which we intend to experiment in the
future.  The marginalisation of the latent space allows for such an
expansion with simultaneous regularisation. 
The variational lower bound allows us to make a principled choice
between models trained using different initializations and with
different numbers of layers.
The deep hierarchy we have proposed can also be used with inputs
governing the top layer of the hierarchy, leading to a powerful model
for regression based on Gaussian processes, but which is not itself a
Gaussian process. In the future, we wish to test this model for applications in
multitask learning (where intermediate layers could learn
representations shared across the tasks) and in modelling
nonstationary data or data involving jumps. These are both areas where
a single layer GP struggles.

A remaining challenge is to extend our methodologies to very large data sets. A very promising approach would be to apply \emph{stochastic variational inference} \citep{Hoffman:stochastic12}. In a recent workshop publication \cite{Hensman:stochastic12} have shown that the standard variational GP and Bayesian GP-LVM can be made to fit within this formalism. The next step for deep GPs will be to incorporate these large scale variational learning algorithms.


\subsubsection*{Acknowledgements}
Research was supported by the University of Sheffield Moody endowment fund and the Greek State
Scholarships Foundation (IKY).  



\bibliographystyle{abbrvnat}
\bibliography{deepGPs,lawrence,other,zbooks}

\end{document}